\documentclass[runningheads]{llncs}
\usepackage{graphicx}
\usepackage{tikz}
\usepackage{comment}
\usepackage{amsmath,amssymb} % define this before the line numbering.
\usepackage{color}
\usepackage{orcidlink}

\usepackage{hyperref}
\usepackage{booktabs}
\usepackage{multirow}
\usepackage{subfigure}
\newcommand{\mname}{TIE}
\usepackage{amsmath,amsfonts,bm}

\def\eqref#1{equation~\ref{#1}}
\def\Eqref#1{Equation~\ref{#1}}
\def\1{\bm{1}}

\newcommand{\test}{\mathcal{D_{\mathrm{test}}}}

\def\va{{\bm{a}}}

\def\ve{{\bm{e}}}

\def\vp{{\bm{p}}}
\def\vq{{\bm{q}}}
\def\vr{{\bm{r}}}
\def\vs{{\bm{s}}}

\def\vv{{\bm{v}}}

\def\vx{{\bm{x}}}

\DeclareMathAlphabet{\mathsfit}{\encodingdefault}{\sfdefault}{m}{sl}
\SetMathAlphabet{\mathsfit}{bold}{\encodingdefault}{\sfdefault}{bx}{n}

\newcommand{\softmax}{\mathrm{softmax}}

\newcommand{\eg}{\emph{e.g.}}

\begin{document}
% \linenumbers
\pagestyle{headings}
\mainmatter
\def\ECCVSubNumber{6639}  % Insert your submission number here

\title{Transformer with Implicit Edges for Particle-based Physics Simulation} % Replace with your title

\begin{comment}
\titlerunning{ECCV-22 submission ID \ECCVSubNumber}
\authorrunning{ECCV-22 submission ID \ECCVSubNumber}
\author{Anonymous ECCV submission}
\institute{Paper ID \ECCVSubNumber}
\end{comment}
%******************

%\begin{comment}
\titlerunning{Transformer with Implicit Edges}
\author{Yidi Shao\inst{1}\orcidlink{0000-0001-8020-5150} \and
Chen Change Loy\inst{1}\orcidlink{0000-0001-5345-1591}\index{Chen Change, Loy} \and
Bo Dai\inst{2}\orcidlink{0000-0003-0777-9232}}
\authorrunning{Y. Shao et al.}
\institute{
    S-Lab for Advanced Intelligence, Nanyang Technological University \\
    \email{yidi001@e.ntu.edu.sg, ccloy@ntu.edu.sg}\\
\and
    Shanghai AI Laboratory \\
    \email{daibo@pjlab.org.cn}\\
}
%\end{comment}
%******************
\maketitle

\begin{abstract}
Particle-based systems provide a flexible and unified way
to simulate physics systems with complex dynamics.
Most existing data-driven simulators for particle-based systems
adopt graph neural networks (GNNs) as their network backbones,
as particles and their interactions can be naturally represented by graph nodes and graph edges.
However, while particle-based systems usually contain hundreds even thousands of particles,
the explicit modeling of particle interactions as graph edges inevitably leads to a significant computational overhead,
due to the increased number of particle interactions.
Consequently,
in this paper we propose a novel Transformer-based method,
dubbed as Transformer with Implicit Edges (\mname),
to capture the rich semantics of particle interactions in an edge-free manner.
The core idea of \mname~is to decentralize the computation involving pair-wise particle interactions
into per-particle updates.
This is achieved by adjusting the self-attention module to resemble the update formula of graph edges in GNN.
To improve the generalization ability of \mname,
we further amend \mname~with learnable material-specific abstract particles
to disentangle global material-wise semantics from local particle-wise semantics.
We evaluate our model on diverse domains of varying complexity and materials.
Compared with existing GNN-based methods,
without bells and whistles,
\mname~achieves superior performance and generalization across all these domains.
Codes and models are available at \url{https://github.com/ftbabi/TIE_ECCV2022.git}\addtocounter{footnote}{-2}\footnote{Bo Dai completed this work when he was with S-Lab, NTU.}.
\end{abstract}

%\thispagestyle{empty}
% !TEX root = ../eccv2022submission.tex

\section{Introduction}
Particle-based physics simulation not only facilitates the exploration of underlying principles in physics, chemistry and biology, it also plays an important role in computer graphics, \eg,~enabling the creation of vivid visual effects such as explosion and fluid dynamic in films and games.
By viewing a system as a composition of particles, particle-based physics simulation imitates system dynamics according to the states of particles as well as their mutual interactions. In this way,
although different systems may contain different materials and follow different physical laws,
they can be simulated in a unified manner with promising quality.

Recent approaches for particle-based physics simulation \cite{battaglia2016interaction,schenck2018spnets,mrowca2018flexible,li2019learning,pmlr-v119-sanchez-gonzalez20a,Ummenhofer2020Lagrangian}
often adopt a graph neural network (GNN) \cite{kipf2016semi} as the backbone network structure,
where particles are treated as graph nodes,
and interactions between neighboring particles are explicitly modeled as edges.
By explicitly modeling particle interactions,
existing methods effectively capture the semantics emerging from those interactions (\eg,~the influences of action-reaction forces),
which are crucial for accurate simulation of complex system.
However,
such an explicit formulation requires the computation of edge features for all valid interactions.
Since a particle-based system usually contains hundreds even thousands of densely distributed particles,
the explicit formulation inevitably leads to significant computational overhead,
limiting the efficiency and scalability of these GNN-based approaches.

\begin{figure}[t]
	\begin{center}
		\includegraphics[width=0.9\textwidth]{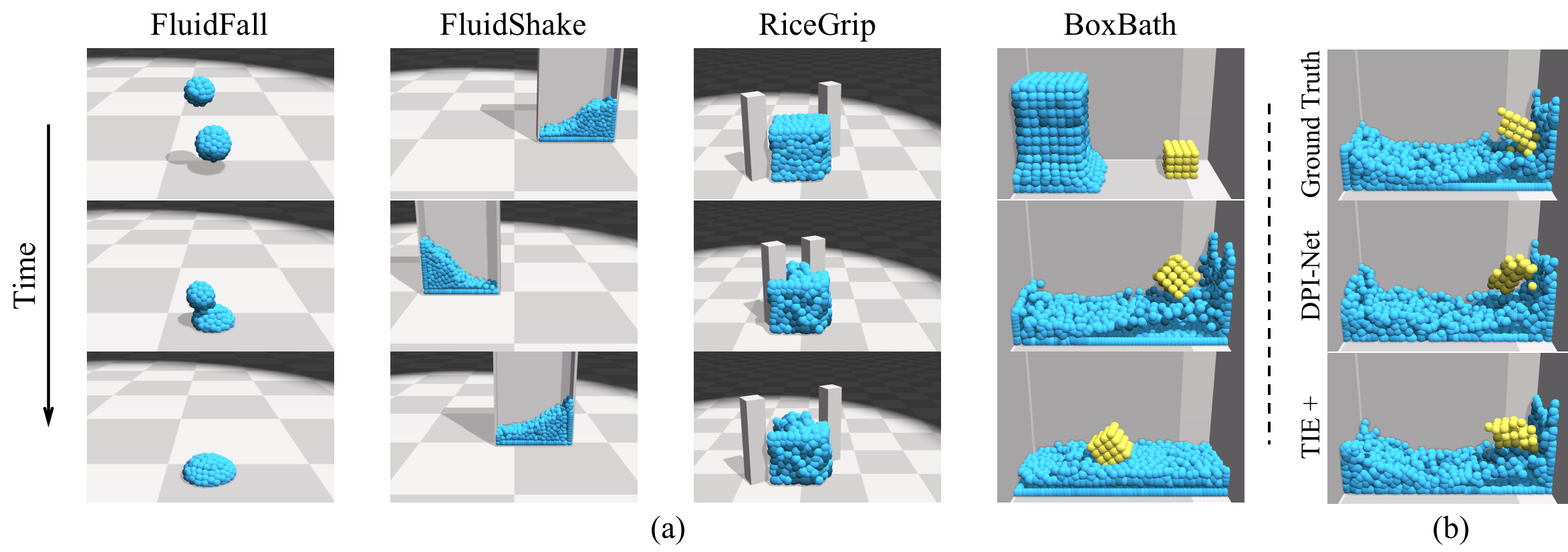}
	\end{center}
	\caption{
	\small{(a). Samples from base domains.
	\emph{FluidFall} contains two drops of water.
	\emph{FluidShake} simulates a block of water in a moving box.
	\emph{RiceGrip} has a deformable object squeezed by two grippers.
	\emph{BoxBath} contains a rigid cubic washed by water.
	(b). Samples from \emph{BunnyBath},
	where we change the rigid cube into bunny for generalization test.
	We compare our \mname~with DPI-Net, which achieves the best performances on \emph{BunnyBath} among previous methods.
	While the bunny is flooded upside down in ground truth,
	\mname~rollouts more faithful results especially in terms of the bunny's posture and fluid dynamics.
	More comparisons can be found in Section \ref{sec:gen}.
	}}
	\label{fig:intro_samples}
\end{figure}

In this paper,
instead of relying on GNN,
we propose to adopt Transformer as the backbone network structure for particle-based physics simulation.
While particle interactions are represented as graph edges in GNN,
in Transformer they are captured by a series of self-attention operations,
in the form of dot-products between tokens of interacting particles.
Consequently,
in Transformer only particle tokens are required to simulate a system,
leading to significantly reduced computational complexity when compared to GNN-based approaches.

The vanilla Transformer, however, is not directly applicable for effective simulation,
since the rich semantics of particle interactions cannot be fully conveyed by the dot-products in self-attention operations.
In this work, we address the problem via a novel modification to the self-attention operation that resembles the effect of edges in GNN
but exempts from the need of explicitly modeling them.
Specifically,
each particle token is decomposed into three tokens,
namely a state token, a receiver token, and a sender token.
In particular, the state token keeps track of the particle state,
the receiver token describes how the particle's state would change,
and the sender token indicates how the particle will affect its interacting neighbors.
By taking receiver tokens and sender tokens as both keys and values,
while state tokens are queries,
the sturcture of edges in GNN can be equally represented by attention module in Transformer.
To further trace the edge semantics in GNN,
motivated by the process of normalizations for edges,
both receiver and sender tokens are first decentralized in our attention module.
Then we recover the standard deviations of edges from receiver and sender tokens,
and apply the recovered scalar values as part of attention scores.
Thus,
the edge features can be effectively revived in Transformer.
Moreover,
to improve the generalization ability of the proposed method,
we further propose to assign a learnable abstract particle for each type of material,
and force particles of the same material to interact with their corresponding abstract particle,
so that global semantics shared by all particles of the same material can be disentangled from local particle-level semantics.

Our method, dubbed as \textbf{T}ransformer with \textbf{I}mplicit \textbf{E}dges for Particle-based Physics Simulation (\mname),
possesses several advantages over previous methods.
First,
thanks to the proposed edge-free design,
\mname~maintains the same level of computational complexity as in the vanilla Transformer,
while combines the advantages of both Transformer and GNN.
\mname~not only inherits the self-attention operation from Transformer
that can naturally attend to essential particles in the dynamically changing system,
\mname~is also capable of extracting rich semantics from particle interactions as GNN,
without suffering from its significant computational overhead.
Besides,
the introduction of learnable abstract particles further boosts the performance of \mname~in terms of generality and accuracy,
by disentangling global semantics such as the intrinsic characteristics of different materials.
For instance, after learning the dynamics of water,
\mname~can be directly applied to systems with varying numbers and configurations of particles,
mimicking various effects including waterfall and flood.

To demonstrate the effectiveness of \mname,
a comprehensive evaluation is conducted on four standard environments
commonly used in the literature \cite{li2019learning,pmlr-v119-sanchez-gonzalez20a},
covering domains of different complexity and materials,
where \mname~achieves superior performance across all these environments compared to existing methods.
Attractive properties of \mname,
such as its strong generalization ability,
are also studied,
where we adjust the number and configuration of particles in each environments to create unseen systems for \mname~to simulate without re-training.
Compared to previous methods,
\mname~is able to obtain more realistic simulation results across most unseen systems.
For example,
after changing the shape from cube to bunny in \emph{BoxBath},
the MSEs achieved by \mname{} is at least 30\% lower than previous methods.

% !TEX root = ../eccv2022submission.tex

\section{Related Work} \label{related_work}

\textbf{Physics simulation by neural networks}.
There are many different kind of representations for physics simulations. 
Grid-based methods \cite{lee2019data,thuerey2020deep,wang2020towards}
adopt convolutional architectures for learning high-dimensional physical system,
while mesh-based simulations \cite{bronstein2017geometric,luo2018nnwarp,Hanocka_2019,nash2020polygen,qiao2020scalable,weng2021graphbased,pfaff2021learning}
typically simulate objects with continuous surfaces, such as clothes, rigid objects, surfaces of water and so on.

Many studies \cite{battaglia2016interaction,schenck2018spnets,mrowca2018flexible,li2019learning,Ummenhofer2020Lagrangian,pmlr-v119-sanchez-gonzalez20a}
simulate physics on particle-based systems,
where all objects are represented by groups of particles.
Specifically,
Interaction Network (IN) \cite{battaglia2016interaction} simulated interactions in object-level.
Smooth Particle Networks (SPNets) \cite{schenck2018spnets}
implemented fluid dynamics using position-based fluids \cite{macklin2013position}.
Hierarchical Relation Network (HRN) \cite{mrowca2018flexible}
predicted physical dynamics based on hierarchical graph convolution.
Dynamic Particle Interaction Networks (DPI-Net) \cite{li2019learning} combined dynamic graphs,
multi-step spatial propagation,
and hierarchical structure to simulate particles.
CConv \cite{Ummenhofer2020Lagrangian} used spatial convolutions to simulate fluid particles.
Graph Network-based Simulators (GNS) \cite{pmlr-v119-sanchez-gonzalez20a} 
computed dynamics via learned message-passing.

Previous work mostly adopted graph networks for simulations.
They extracted potential semantics by explicitly modeling edges and storing their embeddings,
and required each particle to interact with all its nearby particles
without selective mechanism.
In contrast,
our \mname~is able to capture semantics in edges in an edge-free manner,
and selectively focus on necessary particle interactions through attention mechanism.
Experiments show that
\mname~is more efficient,
and surpasses existing GNN-based methods.

\noindent
\textbf{Transformer}.
Transformer \cite{vaswani2017attention} was designed for machine translation
and achieved state-of-the-art performance in many natural langruage processing tasks
\cite{devlin2019bert,radford2019language,brown2020language}.
Recently,
Transformer starts to show great expandability and applicability in many other fields,
such as computer vision \cite{wang2018non,carion2020end,dosovitskiy2021image,wang2021maxdeeplab,liu2021swin},
and graph representations \cite{zhou2020data,zhang2020graph,dwivedi2021generalization}.
To our knowledge,
no attempt has been made to apply Transformer on physics simulation.

Our \mname~inherits the multi-head attention mechanism,
contributing to dynamically model the potential pattern in particle interactions.
Though Graph Transformer \cite{dwivedi2021generalization},
which we refer as GraphTrans for short,
is also Transformer-based model on graphs,
it still turns to explicitly modeling each valid edge to enhance the semantics of particle tokens,
failing to make full use of attention mechanism to describe the relations among tokens in a more efficient manner.
We adopt GraphTrans \cite{dwivedi2021generalization} in particle-based simulation
and compare it with \mname~in experiments.
Quantitative and qualitative results show that
\mname~achieves more faithful rollouts in a more efficient way. 

% !TEX root = ../eccv2022submission.tex

\section{Methodology} \label{method}
\subsection{Problem Formulation}
For a particle-based system composed of $N$ particles,
we use $\mathcal{X}^t = \{ \vx^t_{i}\}_{i=1}^N$ to denote the system state at time step $t$,
where $\vx^t_{i}$ denotes the state of $i$-th particle.
Specifically, $\vx^t_i = [\vp^t_i, \vq^t_i, \va_i]$,
where $\vp^t_i, \vq^t_i \in \mathbb{R}^3$ refer to position and velocity,
and $\va_i \in \mathbb{R}^{d_a}$ represents fixed particle attributes such as its material type.
The goal of a simulator is to learn a model $\phi(\cdot)$ from previous rollouts of a system
to causally predict a rollout trajectory in a specific time period
conditioned on the initial system state $\mathcal{X}^0$.
The prediction runs in a recursive manner,
where the simulator will predict the state $\hat{\mathcal{X}}^{t+1} = \phi(\mathcal{X}^t)$ at time step $t+1$ based on the state $\mathcal{X}^t = \{ x^t_{i}\}$ at time step $t$.
In practice,
we will predict the velocities of particles $\hat{Q}^{t+1}=\{\hat{\vq}^{t+1}_i\}$,
and obtain their positions via
$\hat{\vp}^{t+1}_i = \vp^t_i + \Delta t \cdot \hat{\vq}^{t+1}_i$,
where $\Delta t$ is a domain-specific constant.
In the following discussion,
the time-step $t$ is omitted to avoid verbose notations.

\subsection{GNN-based Approach}
As particle-based physics systems can be naturally viewed as directed graphs,
a straightforward solution for particle-based physics simulation is applying graph neural network (GNN)
\cite{battaglia2016interaction,schenck2018spnets,mrowca2018flexible,li2019learning,pmlr-v119-sanchez-gonzalez20a}.
Specifically,
we can regard particles in the system as graph nodes,
and interactions between pairs of particles as directed edges.
Given the states of particles $\mathcal{X} = \{ \vx_{i}\}_{i=1}^N$ at some time-step,
to predict the velocities of particles in the next time-step,
GNN will at first obtain the initial node features and edge features following:
\begin{align}
\vv_i^{(0)}	&	=	f^{\mathrm{enc}}_V(\vx_i), \label{func:v_enc} \\
\ve_{ij}^{(0)}	&	=	f^{\mathrm{enc}}_E (\vx_i, \vx_j), \label{func:e_enc}
\end{align}
where $\vv_i, \ve_{ij} \in \mathbb{R}^{d_h}$ are $d_h$ dimensional vectors,
and $f^{\mathrm{enc}}_V(\cdot), f^{\mathrm{enc}}_E(\cdot)$ are respectively the node and edge encoders.
Subsequently,
GNN will conduct $L$ rounds of message-passing, and obtain the velocities of particles as:
\begin{align}
\ve^{(l+1)}_{ij}	&	=	f^{\mathrm{prop}}_E (\vv^{(l)}_{i}, \vv^{(l)}_{j}, \ve^{(l)}_{ij}), \label{func:e_prop} \\
\vv^{(l+1)}_{i}	&	=	f^{\mathrm{prop}}_V(\vv^{(l)}_{i}, \sum_{j \in \mathcal{N}_i} \ve^{(l+1)}_{ij}), \label{func:v_prop} \\
\hat{\vq}_i	&	=	f^{\mathrm{dec}}_V(\vv^{(L)}_i),  \label{func:p_pred}
\end{align}
where $\mathcal{N}_i$ indicates the set of neighbors of $i$-th particle,
and $f^{\mathrm{prop}}_E (\cdot), f^{\mathrm{prop}}_V (\cdot)$ and $f^{\mathrm{dec}}_V(\cdot)$ are respectively the node propagation module, the edge propagation module as well as the node decoder.
In practice, $f^{\mathrm{enc}}_V(\cdot), f^{\mathrm{enc}}_E(\cdot), f^{\mathrm{prop}}_E (\cdot), f^{\mathrm{prop}}_V (\cdot)$ and $f^{\mathrm{dec}}_V(\cdot)$ are often implemented as multi-layer perceptrons (MLPs).
Moreover,
a window function $g$ is commonly used to filter out interactions between distant particles and reduce computational complexity:
\begin{align}
g(i,j)	&	=	\mathbf{1}\left(\Vert \vp_i - \vp_j\Vert_2 < R\right), \label{func:window_func}
\end{align}
where $\mathbf{1}(\cdot)$ is the indicator function
and $R$ is a pre-defined threshould.

\begin{figure}[t]
	\begin{center}
		\subfigure[Edge propagations in GNNs.]{\includegraphics[width=120mm]{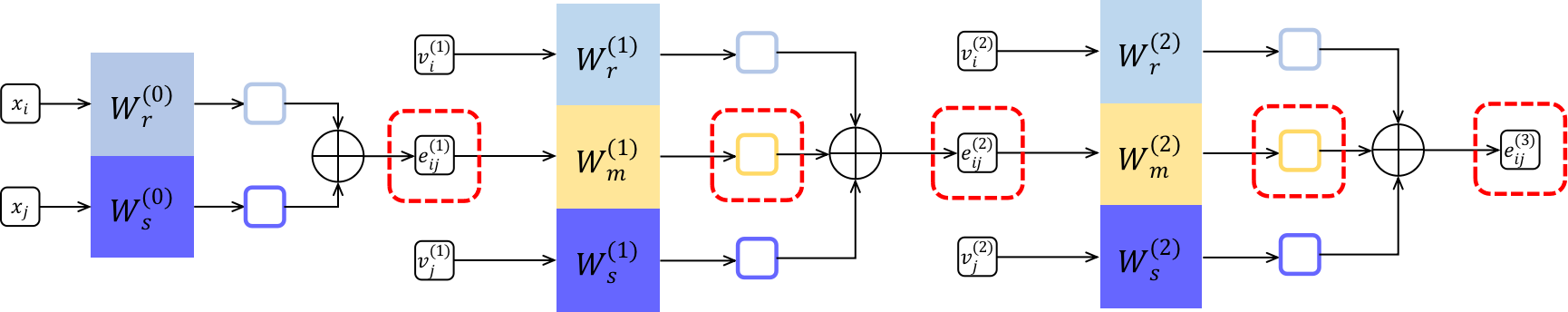}} \label{fig:gnn_path}\\% \label{fig:base}
		\subfigure[Implicit edge propagations in \mname{}.]{\includegraphics[width=120mm]{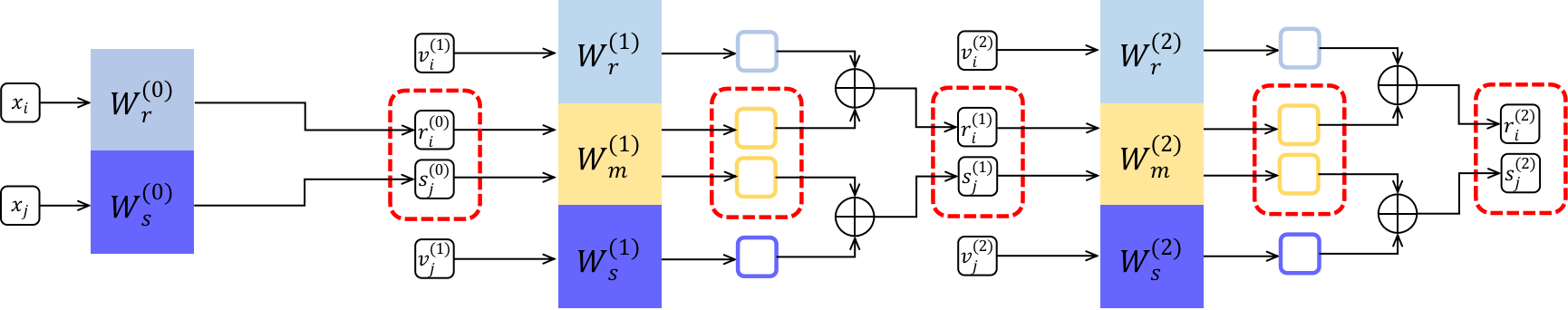}} \label{fig:tiips_path}
	\end{center}
	\caption{\small{
	We demonstrate the propagations for edges in GNNs and \mname{},
	where explicit or implicit edges are shown in red boxes.
	The process of MLP in each layer is splitted into blocks of square followed by summations.
	Different blocks of MLP are shown by square areas with different colors.
	The key idea of \mname{} is that
	\mname{} replaces the explicit edges $e_{ij}^{(l+1)}$ by receiver tokens $r_i^{(l)}$
	and sender tokens $s_j^{(l)}$.
	When only considering the trainable weights of each MLP,
	the summation of receiver and sender tokens within a red box equals to the edge within the same depth of red box, as shown in \Eqref{func:e_rs}.
	From the indexes we can know,
	the behaviors of node $i$ and $j$ are independent in Figure (b),
	thus \mname{} does not include explicit edges.}}
	\label{fig:dec_edge}
\end{figure}

\subsection{From GNN to Transformer}
To accurately simulate the changes of a system over time,
it is crucial to exploit the rich semantics conveyed by the interactions among particles,
such as the energy transition of a system when constrained by material characteristics and physical laws.
While GNN achieves this by explicitly modeling particle interactions as graph edges,
such a treatment also leads to substantial computational overhead.
Since a particle-based system contains hundreds even thousands of particles,
and particles of a system are densely clustered together,
this issue significantly limits the efficiency of GNN-based approaches.

Inspired by recent successes of Transformer \cite{vaswani2017attention}
that applies computational efficient self-attention operations to model the communication among different tokens,
in this paper we propose a Transformer-based method,
which we refer to as Transformer with Implicit Edges, \mname,
for particle-based physics simulation.
We first describe how to apply a vanilla Transformer in this task.
Specifically,
we assign a token to each particle of the system,
and therefore particle interactions are naturally achieved by $L$ blocks of multi-head self-attention modules.
While the token features are initialized according to \Eqref{func:v_enc},
they will be updated in the $l$-th block as:
\begin{eqnarray}
	\omega_{ij}		&	=	&	(W^{(l)}_Q \vv^{(l)}_i)^\top \cdot (W^{(l)}_K \vv^{(l)}_{j}), \\ \label{func:v2v_scaler_vanilla}
	\vv^{(l+1)}_i	&	=	&	\sum_j \frac{\exp(\omega_{ij}{/\sqrt{d}})}{\sum_{k} \exp(\omega_{i{k}}{/\sqrt{d}})} \cdot (W^{(l)}_V\vv^{(l)}_{j}), \label{func:agg}
\end{eqnarray}
where $d$ is the dimension of features, and $W^{(l)}_Q, W^{(l)}_K, W^{(l)}_V$ are weight matrices for queries, keys, and values.
Following the standard practice,
a mask is generated according to \Eqref{func:window_func} to mask out distant particles when computing the attention.
And finally,
the prediction of velocities follows \Eqref{func:p_pred}.

Although the vanilla Transformer provides a flexible approach for particle-based simulation,
directly applying it leads to inferior simulation results as shown in our experiments.
In particular,
the vanilla Transformer uses attention weights that are scalars obtained via dot-product,
to represent particle interactions,
which are insufficient to reflect the rich semantics of particle interactions.
To combine the merits of GNN and Transformer,
\mname~modify the self-attention operation in the vanilla Transformer to
\emph{implicitly} include edge features as in GNN in an edge-free manner.
In Figure \ref{fig:dec_edge} we include the comparison between our proposed implicit edges and the explicit edges in GNN.
Specifically,
since $f^{\mathrm{prop}}_E$ in \Eqref{func:e_prop} and $f^{\mathrm{enc}}_E$ in \Eqref{func:e_enc} are both implemented as an MLP in practice,
by expanding \Eqref{func:e_prop} recursively and grouping terms respectively for $i$-th and $j$-th particle we can obtain:
\begin{align}
	\vr^{(0)}_i & = W^{(0)}_r \vx_i, \qquad \vs^{(0)}_j=W^{(0)}_s \vx_j, \\
	\vr^{(l)}_i & = W^{(l)}_r \vv^{(l)}_i + W^{(l)}_m \vr^{(l-1)}_i, \\
	\vs^{(l)}_j & = W^{(l)}_s \vv^{(l)}_j + W^{(l)}_m \vs^{(l-1)}_j, \\
	\ve^{(l+1)}_{ij} & = \vr^{(l)}_i + \vs^{(l)}_j, \label{func:e_rs}
\end{align}
where the effect of an explicit edge can be effectively achieved by two additional tokens,
which we refer to as the receiver token $\vr_i$ and the sender token $\vs_j$.
The detained expansion of \Eqref{func:e_prop} can be found in the supplemental material.
Following the above expansion,
\mname~thus assigns three tokens to each particle of the system,
namely a receiver token $\vr_i$, a sender token $\vs_i$, and a state token $\vv_i$.
The state token is similar to the particle token in the vanilla Transformer,
and its update formula combines the node update formula in GNN and the self-attention formula in Transformer:
\begin{eqnarray}
	\omega^\prime_{ij}		&	=	&	(W^{(l)}_Q \vv^{(l)}_i)^\top \vr^{(l)}_i + (W^{(l)}_Q \vv^{(l)}_i)^\top \vs^{(l)}_j,  \label{func:v2v_scaler} \\
	\vv^{(l+1)}_i	&	=	&	\vr^{(l)}_i + \sum_j \frac{\exp(\omega^\prime_{ij}{/\sqrt{d}})}{\sum_{k} \exp(\omega^\prime_{i{k}}{/\sqrt{d}})} \cdot \vs^{(l)}_j. \label{func:ve_agg}
\end{eqnarray}
We refer to \Eqref{func:ve_agg} as an implicit way to incorporate the rich semantics of particle interactions,
since \mname~approximates graph edges in GNN with two additional tokens per particle,
and more importantly these two tokens can be updated, along with the original token, separately for each particle,
avoiding the significant computational overhead.
To interpret the modified self-attention in \mname,
from the perspective of graph edges,
we decompose them into the receiver tokens and the sender tokens,
maintaining two extra paths in the Transformer's self-attention module.
As for the perspective of self-attention,
the receiver tokens and the sender tokens respectively replace the original keys and values.

In practice, since GNN-based methods usually incorporate LayerNorm \cite{DBLP:journals/corr/BaKH16} in their network architectures that computes the mean and std of edge features to improve their performance and training speed,
we can further modify the self-attention in \Eqref{func:v2v_scaler} and \Eqref{func:ve_agg} to include the effect of normalization as well:
\begin{eqnarray}
	{\left(\sigma^{(l)}_{ij}\right)^2}	&	=	&	\frac{1}{d} (\vr^{(l)}_{i})^\top\vr^{(l)}_{i} + \frac{1}{d} (\vs^{(l)}_{j})^\top\vs^{(l)}_{j} + \frac{2}{d}(\vr^{(l)}_{i})^\top \vs^{(l)}_{j} -(\mu^{(l)}_{{r_i}}+\mu^{(l)}_{{s_j}})^2, \label{func:std_rs} \\
	\omega^{\prime\prime}_{ij}	&	=	&	\frac{(W^{(l)}_Q\vv^{(l)}_i)^\top (\vr^{(l)}_i-\mu^{(l)}_{{r_i}}) + (W^{(l)}_Q \vv^{(l)}_i)^\top (\vs^{(l)}_j-\mu^{(l)}_{{s_j}})}{{\sigma^{(l)}_{ij}}},\label{attn_norm} \\
	\vv^{(l+1)}_i	&	=	&	\sum_j \frac{\exp(\omega^{\prime\prime}_{ij}{/\sqrt{d}})}{\sum_{k} \exp(\omega^{\prime\prime}_{i{k}}{/\sqrt{d}})} \cdot {\frac{(\vr^{(l)}_i-\mu^{(l)}_{r_i}) + (\vs^{(l)}_j - \mu^{(l)}_{s_j})}{\sigma^{(l)}_{ij}}}, \label{func:norm_agg}
\end{eqnarray}
where $\mu^{(l)}_{{r_i}}$ and $\mu^{(l)}_{{s_j}}$ are respectively the mean of receiver tokens and sender tokens after $l$-th block.
Detailed deduction can be found in the supplemental material.

\subsection{Abstract Particles}
To further improve the generalization ability of \mname~and disentangle global material-specific semantics from local particle-wise semantics,
we further equip \mname~with material-specific abstract particles.

For $N_a$ types of materials,
\mname~respectively adopts $N_a$ abstract particles $A=\{\va_k\}_{k=1}^{N_a}$,
each of which is a virtual particle with a learnable state token.
Ideally, the abstract particle $\va_k$ should capture the material-specific semantics of $k$-th material.
They act as additional particles in the system,
and their update formulas are the same as normal particles.
Unlike normal particles that only interact with neighboring particles,
each abstract particle is forced to interact with all particles belonging to its corresponding material.
Therefore with $N_a$ abstract particles
\mname~will have $N+N_a$ particles in total:
$\{ \va_1, \cdots, \va_{N_a} , \vx_1, \cdots, \vx_{N}\}$.
Once \mname~is trained,
abstract particles can be reused when generalizing \mname~to unseen domains that have same materials but vary in particle amont and configuration.

\subsection{Traning Objective and Evaluation Metric}
To train \mname{} with existing rollouts of a domain,
the standard mean square error (MSE) loss is applied to the output of \mname:
\begin{eqnarray}
	\text{MSE}(\hat{Q}, Q)	&	=	&	\frac{1}{N}\sum_i \Vert \hat{\vq}_i-\vq_i \Vert_2^2  \label{func:mse},
\end{eqnarray}
where $\hat{Q} = \{\hat{\vq}_i\}_{i=1}^N$ and $Q = \{\vq_i\}_{i=1}^N$ are respectively the estimation and the ground truth,
and $\Vert \cdot \Vert_2$ is the L2 norm.

In terms of evaluation metric,
since a system usually contains multiple types of materials with imbalanced numbers of particles,
to better reflect the estimation accuracy,
we apply the Mean of Material-wise MSE ($\text{M}^3\text{SE}$) for evaluation:
\begin{eqnarray}
	\text{M}^3\text{SE}(\hat{Q}, Q)	&	=	&	\frac{1}{K}\sum_k\frac{1}{N_k}\sum_i\Vert\hat{\vq}_{i,k}-\vq_{i,k}\Vert_2^2   \label{func:wmse},
\end{eqnarray}
where $K$ is the number of material types,
$N_k$ is the number of particles belonging to the $k$-th material.
$\text{M}^3\text{SE}$ is equivalent to the standard MSE when $K=1$.

% !TEX root = ../eccv2022submission.tex

\begin{table}[t!]
	\caption{\small{
	We report M$^3$SEs (1e-2) results on four base domains,
	while keep the models' number of parameters similar to each other.
	\mname{} achieves superior performance on all domains without suffering from its
	significant computational overhead.
	When adding trainable abstract particles,
	\mname{}, marked by +, further improves performance on \emph{RiceGrip} and \emph{BoxBath},
	which involve complex deformations and multi-material interactions respectively.}}
	\label{tbl:quantitative}
	\setlength{\tabcolsep}{2.3pt}
	\begin{center}
	\scriptsize
	\begin{tabular}{p{1.9cm}cccccccc}
	\toprule
	\multirow{2}{*}{\bf Methods}	& \multicolumn{2}{c}{\bf FluidFall} 	& \multicolumn{2}{c}{\bf FluidShake} & \multicolumn{2}{c}{\bf RiceGrip}	& \multicolumn{2}{c}{\bf BoxBath} \\ \cmidrule(lr{0.75em}){2-3} \cmidrule(lr{0.75em}){4-5} \cmidrule(lr{0.75em}){6-7} \cmidrule(lr{0.75em}){8-9}
	& \bf M$^3$SE	& \bf \#Para & \bf M$^3$SE	& \bf \#Para 	& \bf M$^3$SE	& \bf \#Para & \bf M$^3$SE	& \bf \#Para
	\\ \midrule
	DPI-Net \cite{li2019learning}	&0.08$\pm$0.05	&0.61M	&1.38$\pm$0.45	&0.62M	&0.13$\pm$0.09		&1.98M	 &1.33$\pm$0.29		&1.98M\\
	CConv \cite{Ummenhofer2020Lagrangian}		&0.08$\pm$0.02	&0.84M	&1.41$\pm$0.46	&0.84M	& N/A & N/A & N/A & N/A\\
	GNS	\cite{pmlr-v119-sanchez-gonzalez20a}		&0.09$\pm$0.02	&0.70M	&1.66$\pm$0.37	&0.70M	&0.40$\pm$0.16	&0.71M	& 1.56$\pm$0.23	&0.70M\\
	GraphTrans \cite{dwivedi2021generalization}	&0.04$\pm$0.01	&0.77M	&1.36$\pm$0.37	&0.77M	&0.12$\pm$0.11		&0.78M	&1.27$\pm$0.25		&0.77M\\
	\midrule
	\mname{} (Ours)		&0.04$\pm$0.01		&0.77M	&\textbf{1.22$\pm$0.37}	&0.77M	&0.13$\pm$0.12		&0.78M	&1.35$\pm$0.35		&0.77M\\
	\mname{}+ (Ours)				&\textbf{0.04$\pm$0.00}	&0.77M	&1.30$\pm$0.41	&0.77M	& \textbf{0.08$\pm$0.08}	&0.78M &\textbf{0.92$\pm$0.16}	&0.77M\\
	\bottomrule
	\end{tabular}
	\end{center}
\end{table}
\begin{figure}[t]
	\begin{center}
		\subfigure[\small Batch size is 1.]{\includegraphics[width=27mm]{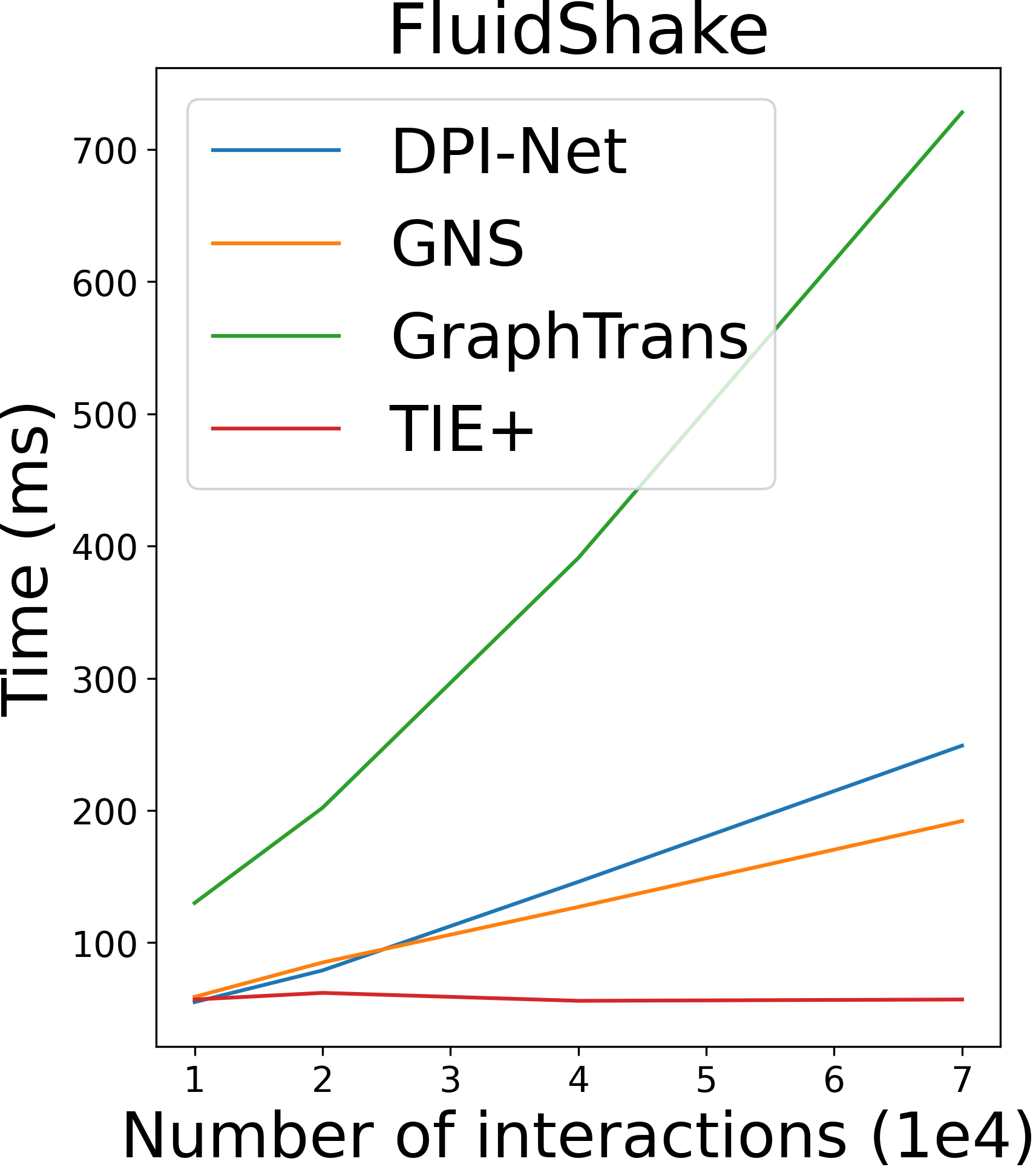}}
		\subfigure[\small Batch size is 1.]{\includegraphics[width=27mm]{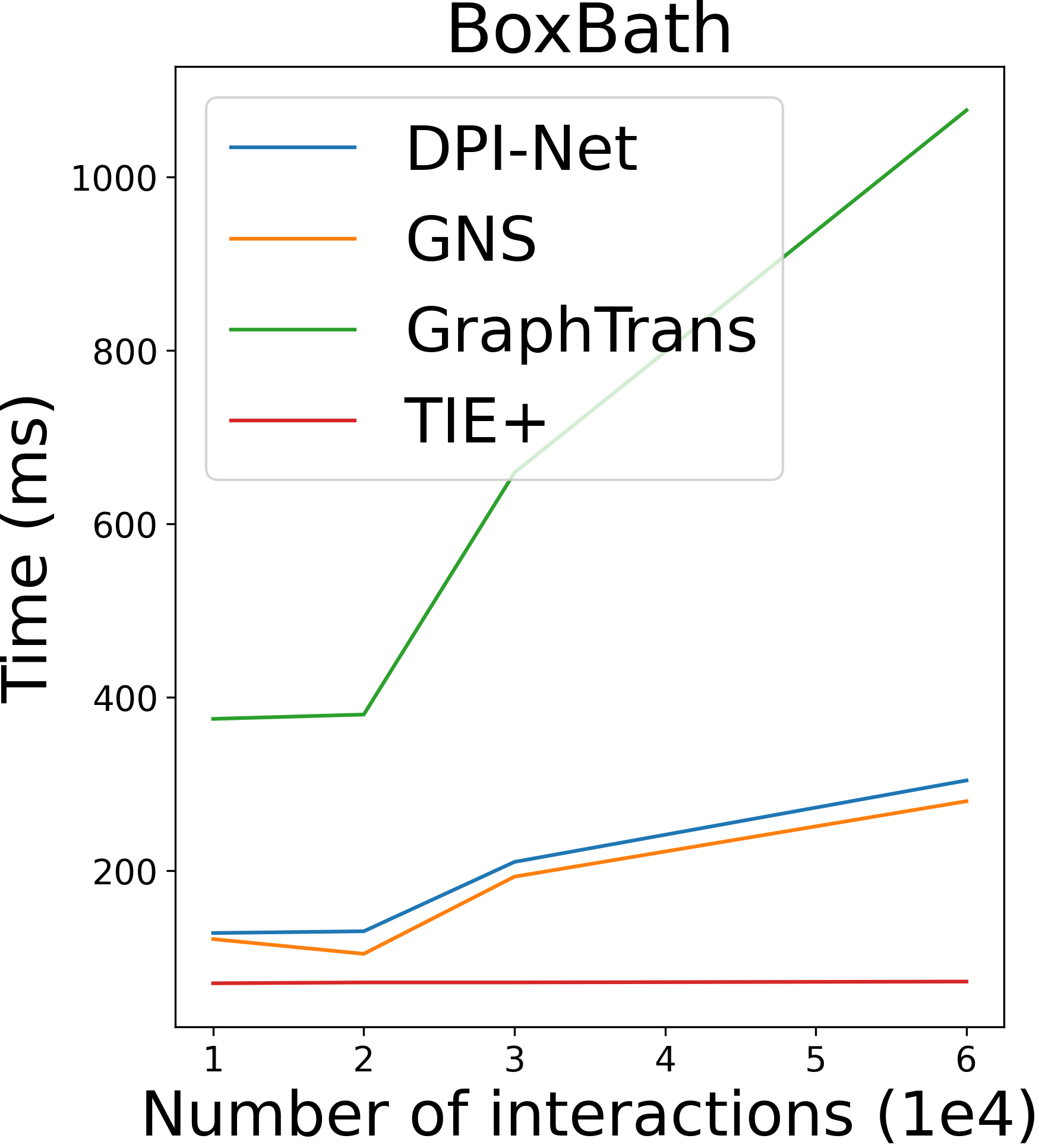}}
		\subfigure[\small Batch size is 4.]{\includegraphics[width=27mm]{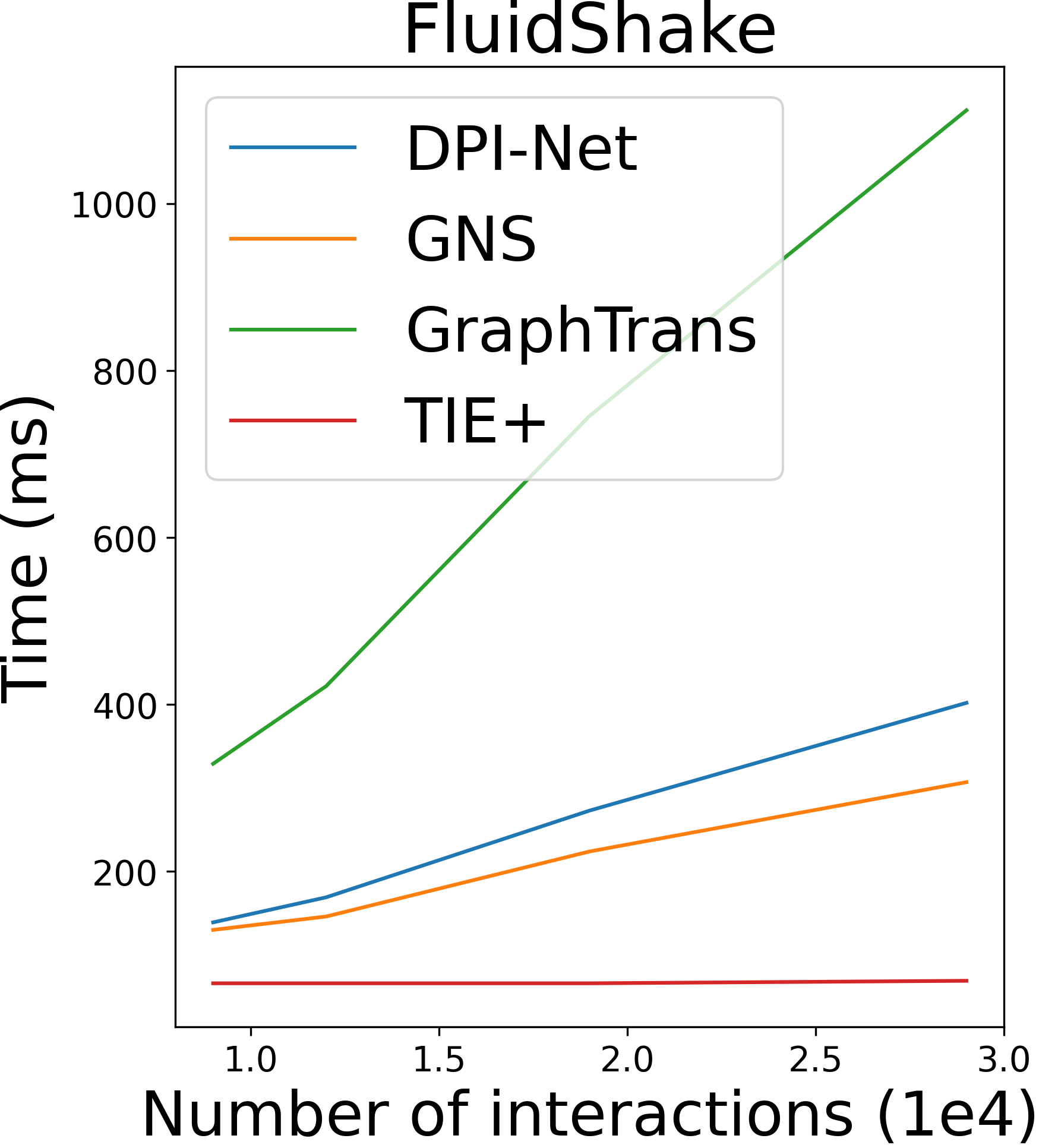}}
		\subfigure[\small Batch size is 4.]{\includegraphics[width=27mm]{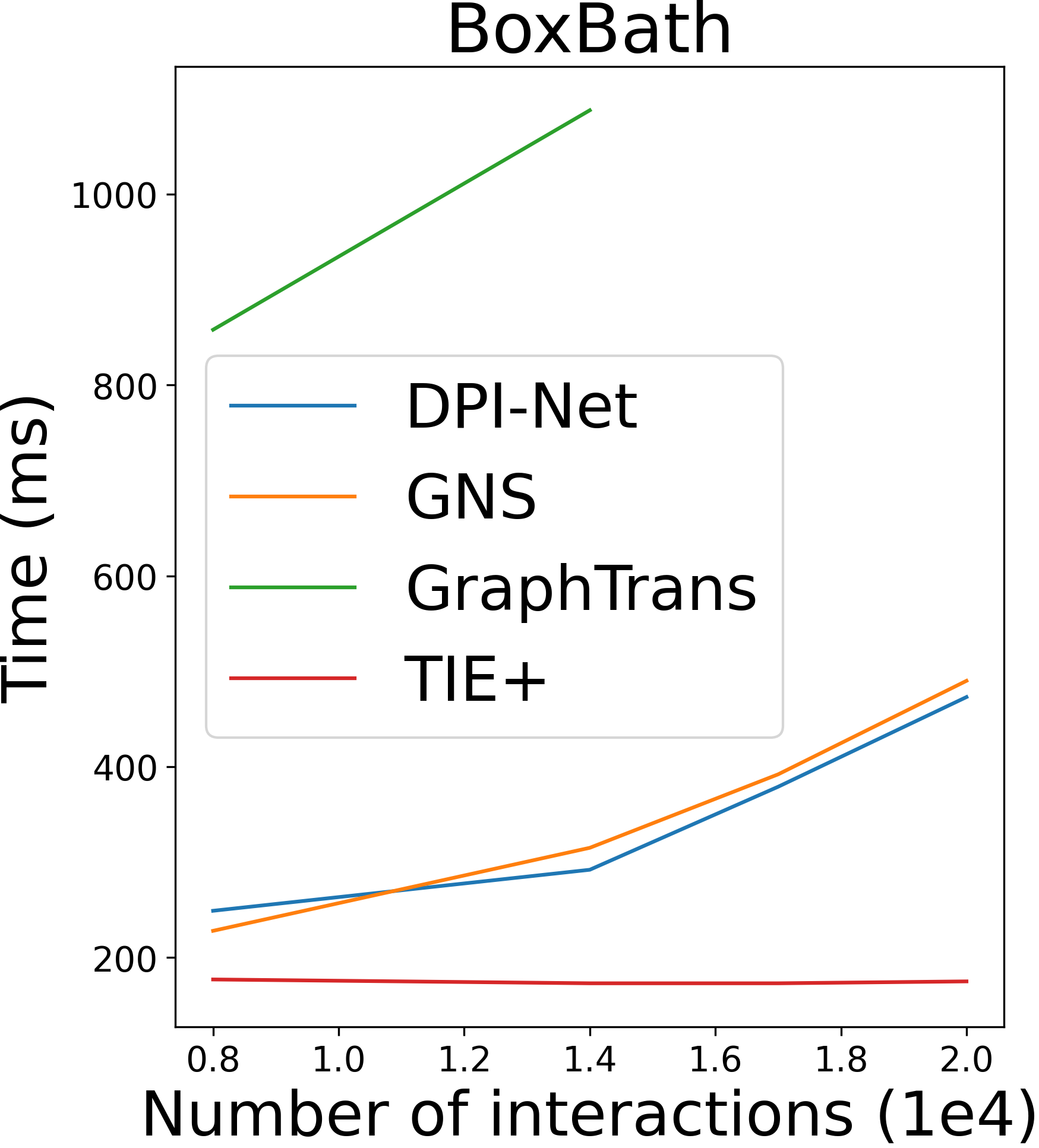}}
		\caption{
		\small{
		We report averaged models' training time for each iteration.
		The batch size in (a) and (b) is set to 1,
		while the batch size in (c) and (d) is set to 4.
		As the number of interactions increases,
		the time cost for \mname+ remains stable,
		while other models spend more time to train due to the computational overhead introduced by extra interactions.
		}}
		\label{fig:time}
	\end{center}
\end{figure}

\section{Experiments} \label{experiment}
We adopt four domains commonly used in the literature
\cite{li2019learning,pmlr-v119-sanchez-gonzalez20a,Ummenhofer2020Lagrangian} for evaluation.
\emph{FluidFall} is a basic simulation for two droplets of water with 189 particle in total;
\emph{FluidShake} is more complex and simulate the water in a randomly moving box,
containing 450 to 627 fluid particles;
\emph{BoxBath} simulates the water washing a rigid cube in fixed box
with 960 fluid particles and 64 rigid particles;
\emph{RiceGrip} simulates the interactions between deformable rice and two rigid grippers,
including 570 to 980 particles.
Samples are displayed in Figure \ref{fig:intro_samples}.
To explore the effectiveness of our model,
we compare \mname{} with four representative approaches:
DPI-Net \cite{li2019learning},
CConv \cite{Ummenhofer2020Lagrangian},
GNS \cite{pmlr-v119-sanchez-gonzalez20a},
and GraphTrans \cite{dwivedi2021generalization}.

\noindent
\textbf{Implementation Details.}
\mname~contains $L=4$ blocks.
For multi-head self-attention version,
The receiver tokens and sender tokens are regarded as the projected keys and values
for each head.
After projecting the concatenated state tokens from all heads,
a two-layer MLPs is followed with dimensions 256 and 128.
The concatenated receiver tokens and sender tokens are directly projected
by one layer MLP with dimensions 128.
The rest hidden dimensions are 128 for default.
We train four models independently on four domains,
with 5 epochs on \emph{FluidShake} and \emph{BoxBath},
13 epochs on \emph{FluidFall},
and 20 epochs on \emph{RiceGrip}.
On \emph{BoxBath},
all models adopt the same strategy to keep the shape of the rigid object following \cite{li2019learning}.
We adopt MSE on velocities as training loss for all models.
The neighborhood radius $R$ in \Eqref{func:window_func} is set to 0.08.
We adopt Adam optimizer with an initial learning rate of 0.0008,
which has a decreasing factor of 0.8 when the validation loss stops to decrease after 3 epochs.
The batch size is set to 16 on all domains.
All models are trained and tested on V100 for all experiments,
with no augmentation involved.

\subsection{Basic Domains}

\begin{figure}[t]
	\begin{center}
		\includegraphics[width=120mm]{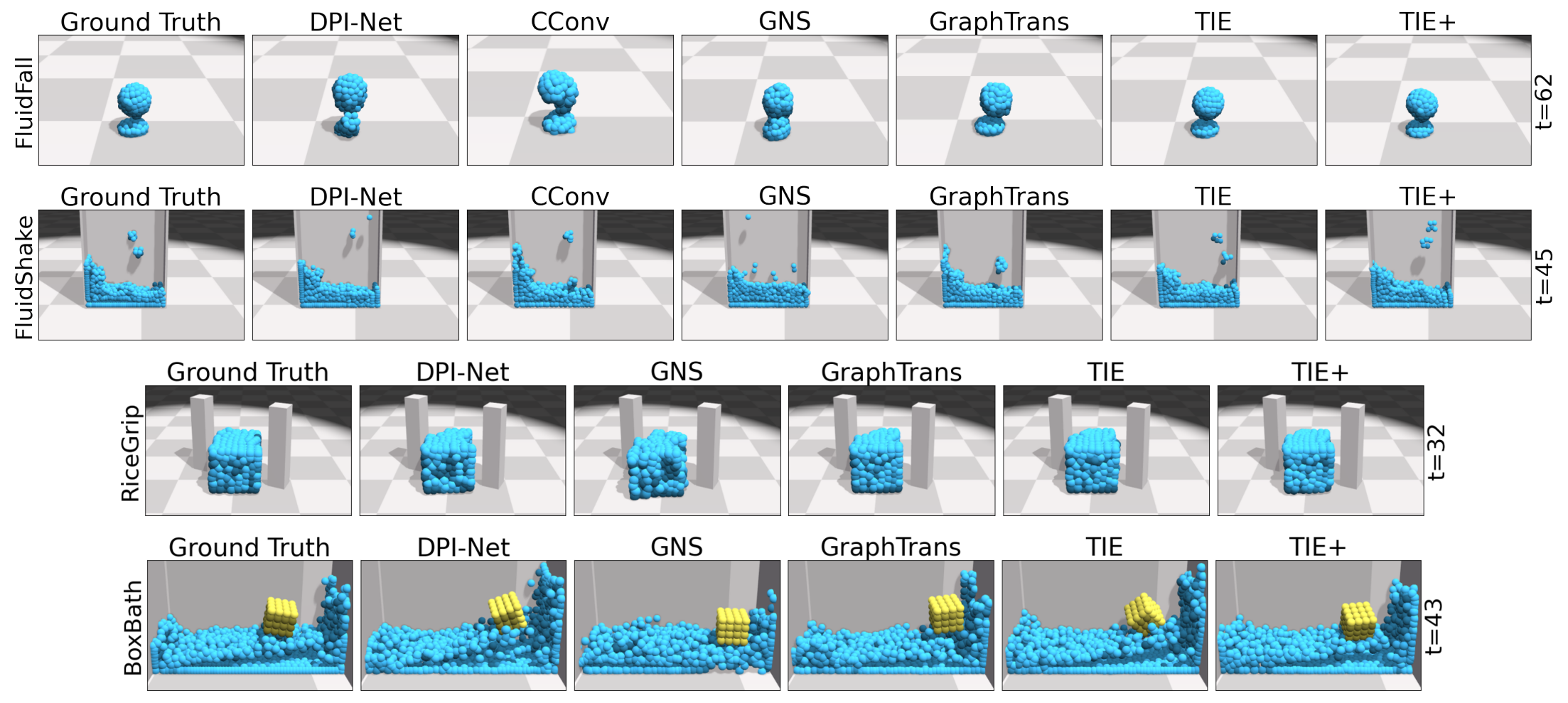}
		\caption{
		\small{
		Qualitative results on base domains.
		\mname{} is able to achieve more faithful results on all domains.
		On \emph{FluidFall},
		\mname{} is able to better maintain the shape before the droplets merge and handle redundant neighbors,
		which are introduced from two different droplets when they move closer to each other.
		The relative positions of the droplets are also closer the the ground truth.
		On \emph{FluidShake},
		\mname{} can predict two faithful blocks of water on the top right.
		On \emph{RiceGrip},
		when focusing on the areas compressed by the grippers,
		the rice restores its position more faithfully in \mname{}.
		On \emph{BoxBath},
		the rigid cube predicted by \mname{} is pushed far away enough from the right wall,
		and the positions for the cube by \mname+ is much closer to the ground truth.
		The fluid particles predicted by our models are also more faithful.
		}}
		\label{fig:base_env}
	\end{center}
\end{figure}

Quantitative results are provided in Table \ref{tbl:quantitative},
while qualitative results are shown in Figure \ref{fig:base_env}.
\mname{} achieves superior performances on all domains.
The effectiveness of abstract particles are more obvious for \emph{RiceGrip} and \emph{BoxBath},
which involve complex materials or multi-material interactions.

\noindent
\textbf{Performance comparison}.
We compare \mname{} with four representative approaches:
DPI-Net \cite{li2019learning},
CConv \cite{Ummenhofer2020Lagrangian},
GNS \cite{pmlr-v119-sanchez-gonzalez20a},
and GraphTrans \cite{dwivedi2021generalization}.
Since DPI-Net and GNS adopt message-passing graph networks for particle-based simulations,
we set the number of propagation steps as four for both models,
except that DPI-Net adopts a total number of six propagation steps on \emph{BoxBath} and \emph{RiceGrip},
where hierarchical structures are adopted.
For CConv,
which designs convolutional layers
carefully tailored to modeling fluid dynamics,
such as an SPH-like local kernel~\cite{monaghan1992smoothed},
we only report the results on fluid-based domains.
As shown in Table~\ref{tbl:quantitative},
\mname{} achieves superior performances on most domains,
while \mname+,
which has abstract particles,
further improves the performances especially
on \emph{RiceGrip} and \emph{BoxBath},
suggesting the effectiveness of abstract particles in modeling complex deformations
and multi-materials interactions.
For qualitative results in Figure \ref{fig:base_env},
our model can predict more faithful rollouts on all domains.

\noindent
\textbf{Efficiency comparison.}
The training time for models with varying batch size is shown in Figure \ref{fig:time}.
For simplicity,
the number of particles is fixed and only number of interactions varies in Figure \ref{fig:time}.
Since \mname~uses implicit edges to model particle interactions and significantly reduces computational overhead,
\mname~has the fastest training speed against general GNN-based simulators (CConv \cite{Ummenhofer2020Lagrangian} is a specialized simulator for systems containing only fluid)
as shown in Figure \ref{fig:time} (a) and (b).
When it comes to batch size larger than 1,
GNN-based methods need pad edges for each batch,
leading to extra computational cost.
On the other hand,
\mname~only need the corresponding attention masks to denote the connectivities without further paddings,
which is faster to train on large batch size.
We does not report the speed of GraphTrans with more than $1.4\times10^4$ interactions due to the limit of memory.
In terms of testing speed,
it is hard to compare different methods directly
since different simulation results will lead to different amount of valid particle interactions.

\subsection{Generalizations}\label{sec:gen}
\begin{table}[t]
\setlength{\tabcolsep}{4pt}
\caption{
\small{
M$^3$SEs on generalizations.
The lists of numbers in \emph{FluidShake} and \emph{RiceGrip} are the range of particles,
while the tuples in \emph{BoxBath} denotes number of fluid particles, number of rigid particles, and shape of rigid objects respectively.
Training settings are marked by *.
\mname{} + achieves the best results on most cases.
	}}
\label{tbl:gen_quantitative}
\begin{center} \scriptsize
\begin{tabular}{lcccc}
\toprule
\multirow{2}{*}{\bf Methods}	&\multicolumn{2}{c}{\bf{FluidShake} [450,627]*}				& \multicolumn{2}{c}{\bf RiceGrip [570,980]*} \\ \cmidrule(lr{.75em}){2-3}\cmidrule(lr{.75em}){4-5}
						&[720,1013]		&[1025,1368] 		&[1062,1347]		&[1349,1642]
\\ \midrule
DPI-Net \cite{li2019learning}        		& 2.13$\pm$0.55	& 2.78$\pm$0.84 	& 0.23$\pm$0.13	& 0.38$\pm$0.67 \\
CConv	\cite{Ummenhofer2020Lagrangian}				& 2.01$\pm$0.55	& \textbf{2.43$\pm$0.81}		& N/A			& N/A    	\\
GNS	\cite{pmlr-v119-sanchez-gonzalez20a}					& 2.61$\pm$0.44	& 3.41$\pm$0.59		& 0.47$\pm$0.20	& 0.51$\pm$0.28 \\
GraphTrans \cite{dwivedi2021generalization}		& 2.68$\pm$0.52	& 3.97$\pm$0.70		& 0.20$\pm$0.13	& 0.22$\pm$0.18 \\
\midrule
\mname{}+ (Ours)		& \textbf{1.92$\pm$0.47}	& 2.46$\pm$0.65			& \textbf{0.17$\pm$0.11} 	& \textbf{0.19$\pm$0.15}\\
\midrule
\multirow{2}{*}{\bf{Methods}}			&\multicolumn{4}{c}{\bf{BoxBath} (960,64,cube)*} \\
\cmidrule(lr{.75em}){2-5}
						&(1280,64,cube) &(960,125,cube)	&(960,136,ball)	&(960,41,bunny)\\
\midrule
DPI-Net	\cite{li2019learning}				& 1.70$\pm$0.22	& 3.22$\pm$0.88	& 2.86$\pm$0.99		& 2.04$\pm$0.79 \\
GNS	\cite{pmlr-v119-sanchez-gonzalez20a}					& 2.97$\pm$0.48	& 2.97$\pm$0.71	& 3.50$\pm$0.67		& 2.17$\pm$0.37 \\
GraphTrans	\cite{dwivedi2021generalization}			& 1.88$\pm$0.25	& 1.50$\pm$0.30	& 1.71$\pm$0.34		& 2.22$\pm$0.61 \\ \midrule
\mname{}+ (Ours)		& \textbf{1.57$\pm$0.18}	& \textbf{1.49$\pm$0.19}	& \textbf{1.45$\pm$0.27}	& \textbf{1.39$\pm$0.48}\\
\bottomrule
\end{tabular}
\end{center}
\end{table}

\begin{figure}[t]
	\begin{center}
		\includegraphics[width=100mm]{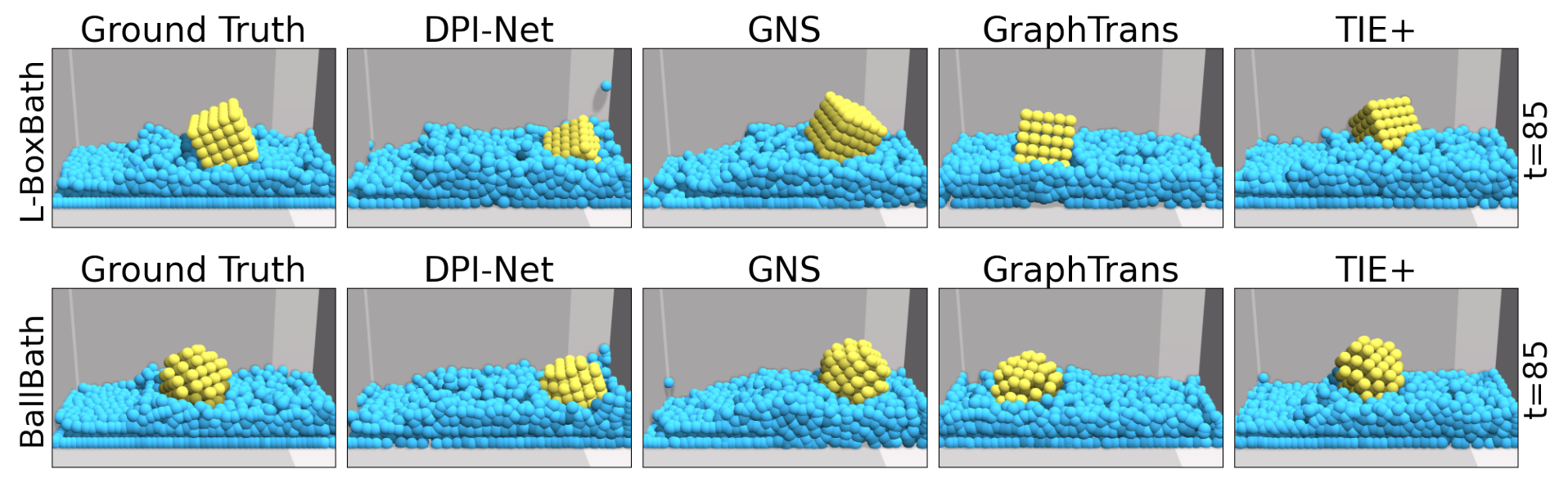}
		\caption{
		\small{
		Qualitative results on generalized domains.
		Here we only show part of results on generalized \emph{BoxBath},
		where we mainly change the shape and size of the rigid object.
		\mname+ can predict more faithful movements of the rigid object,
		while the fluid particles are also vivid.
		More details can be found in supplementary materials.
		}}
		\label{fig:gen_env}
	\end{center}
\end{figure}

As shown in Table~\ref{tbl:gen_quantitative},
we generate more complex domains to challenge the robustness of our full model \mname+.
Specifically,
we add more particles for \emph{FluidShake} and \emph{RiceGrip},
which we refer to as \emph{L-FluidShake} and \emph{L-RiceGrip} respectively.
The \emph{L-FluidShake} includes 720 to 1368 particles,
while \emph{L-RiceGrip} contains 1062 to 1642 particles.
On \emph{BoxBath},
we change the size and shape of rigid object.
Specifically,
we add more fluid particles in \emph{Lfluid-BoxBath} to 1280 fluid particles,
while we enlarge the rigid cube in \emph{L-BoxBath} to 125 particles.
We also change the shape of the rigid object into
ball and bunny,
which we refer to \emph{BallBox} and \emph{BunnyBath} respectively.
Details of generalized environments settings and results can be found in supplementary materials.

Quantitative results are summarized in Table \ref{tbl:gen_quantitative},
while qualitative results are depicted in Figure \ref{fig:gen_env}.
As shown in Table \ref{tbl:gen_quantitative},
\mname+ achieves lower M$^3$SEs on most domains,
while having more faithful rollouts in Figure \ref{fig:gen_env}.
On \emph{L-FluidShake},
\mname+ maintains the block of fluid in the air and predicts faithful wave on the surface.
On \emph{L-RiceGrip},
while DPI-Net and GNS have difficulties in maintaining the shape,
the rice predicted by GraphTrans is compressed more densely only in the center areas
where the grips have reached,
the left side and right side of the rice does not deform properly compared with the ground truth.
In contrast,
\mname+ is able to maintains the shape of the large rice
and faithfully deform the whole rice after compressed.
On generalized \emph{BoxBath},
\mname+ is able to predict faithful rollout
when the fluid particles flood the rigid objects into the air
or when the wave of the fluid particles starts to push the rigid object after the collision.
Even when the rigid object changes to bunny with more complex surfaces,
\mname+ generates more accurate predictions for both fluid particles and rigid particles.

\subsection{Ablation Studies}
We comprehensively analyze our \mname{} and explore the effectiveness of our model in the following aspects:
(a) with and without implicit edges;
(b) with and without normalization effects in attention;
(c) with and without abstract particles;
and (d) the sensitiveness to $R$.
The experiments for (a), (b), and (d) are conducted on \emph{FluidShake} and \emph{L-FluidShake},
while experiment (c) is conducted on \emph{BoxBath}.
The quantitative results are in Table~\ref{tbl:abl} and Table~\ref{tbl:eabl}.

\begin{table}[t]
	\caption{
	\small{Ablation studies. We comprehensively explore the effectiveness of \mname{},
	including the effectiveness of implicitly modeling of edges, normalization effects in attention,
	and abstract particles.
	We report M$^3$SEs(1e-2) on \emph{FluidShake} and \emph{L-FluidShake},
	which are complex domains involving outer forces.}}
	\label{tbl:abl}
	\setlength{\tabcolsep}{2.0pt}
	\begin{center}
	\scriptsize
	\begin{tabular}{lcccc}
	\toprule
	\bf{Configurations}			&A(Transformer)	& B				&C(\mname{})	& D(\mname{}+)
	\\ \midrule
	\bf{Implicit Edges}	&				&\checkmark		&\checkmark		&\checkmark	\\
	\bf{Normalization}	& 				&				&\checkmark			&\checkmark \\
	\bf{Abstract Particles}		& 				&				&				&\checkmark	\\ \midrule
	\bf {FluidShake}				&2.75$\pm$0.86	&1.52$\pm$0.39	&1.22$\pm$0.37 	&1.30$\pm$0.41 \\
	\bf {L-FluidShake}		&8.18$\pm$3.15	&3.17$\pm$0.94	&2.40$\pm$0.74	&2.16$\pm$0.62 \\
	\bottomrule
	\end{tabular}
	\end{center}
\end{table}

\noindent
\textbf{Effectiveness of implicit edges}.
We apply vanilla Transformer encoders by configuration A,
while \mname{} in configuration B does not adopt the interaction attention,
making sure the only difference is the edge-free structure.
The hidden dimension and number of blocks are the same,
while \mname{} is a little larger because of the extra projections for receiver and sender tokens.
As shown in Table~\ref{tbl:abl},
the original Transformer achieves worse performances,
suggesting the scalar attention scores alone are insufficient to capture
rich semantics of interactions among particles.
In contrast,
implicit way of modeling edges enables \mname{} to take advantages of GNN methods
and recover more semantics of particle interactions.

\noindent
\textbf{Effectiveness of normalization effects in attention}.
We follow configuration C to build \mname{},
which includes \Eqref{func:norm_agg}.
Comparing configuration B and C in Table \ref{tbl:abl},
we find that the normalization effects brings benefits to \mname{}
on both base and generalized domains.
Such structure further enables \mname{} to trace the rice semantics from edges,
leading to more stable and robust performances.

\noindent
\textbf{Effectiveness of abstract particles}.
As shown in Table \ref{tbl:eabl},
we replace the abstract particles with dummy particles,
which are zero initialized vectors with fixed values
but have the same connectivities as abstract particles.
Thus,
the dummy particles could not capture the semantics of materials during training.
\mname{} with dummy particles slightly improve the performances on base domains,
suggesting that the extra connectivities introduced by abstract particles
benefit little on \mname{}.
\mname{}+ achieves more stable and robust performances,
suggesting that the abstract particles are able to effectively disentangle the domain-specific semantics,
i.e., the outer forces introduced by walls,
and materials-specific semantics,
i.e., the pattern of fluid particle dynamics.

\begin{table}[t]
	\caption{
	\small{{Ablation studies on abstract particles and sensitiveness to radius $R$.
	To explore the material-aware semantics extracted by abstract particles,
	we conduct experiments on \emph{BoxBath} and the generalized domains \emph{BunnyBath},
	where the rigid cube is replaced by bunny.
	We replace abstract particles with dummy particles,
	which are zero constant vectors and have same connectivities as abstract particles.
	\mname{} marked by "dummy" adopts dummy particles.
	The sensitiveness is on the right part.
	We report M$^3$SEs(1e-2) on \emph{FluidShake}.
	Our default setting on all domains is marked by $*$.}}}
	\label{tbl:eabl}
	\begin{center}
	\scriptsize
	\begin{tabular}{lccclccc}
	\toprule
	\multirow{2}{*}{\bf Methods}	& \multicolumn{2}{c}{\bf BoxBath}	&& \multirow{2}{*}{\bf Methods}	&\multicolumn{3}{c}{\bf{FluidShake}}
	\\ \cmidrule(lr{.75em}){2-3}\cmidrule(lr{.75em}){6-8}
									&(960,64,cube)*	&(960,41,bunny)	&	&& \bf{$R=0.07$}	&\bf{$R^*=0.08$}	&\bf{$R=0.09$}
	\\ \midrule
	\mname{}	&1.35$\pm$0.35	&1.50$\pm$0.45	&&DPI-Net			&2.60$\pm$0.56		&1.38$\pm$0.45			& 1.66$\pm$0.48	\\
	\mname{} dummy	&1.21$\pm$0.28	&1.96$\pm$0.71	&&GraphTrans		&1.97$\pm$0.48					&1.36$\pm$0.37						&1.36$\pm$0.38				\\
	\mname{}+	&0.92$\pm$0.16	&1.39$\pm$0.48	&&\mname{}		&1.60$\pm$0.37		&1.22$\pm$0.37			& 1.31$\pm$0.40 \\
	\bottomrule
	\end{tabular}
	\end{center}
\end{table}

\noindent
\textbf{Sensitiveness to $R$}.
Quantitative results are reported on \emph{FluidShake}.
As shown in Table \ref{tbl:eabl},
when $R$ is smaller,
models tend to have a drop in accuracies due to the insufficient particle interactions.
When $R$ is greater,
the drop of accuracies for DPI-Net is caused
by redundant interactions due to the high flexibility of fluid moving patterns.
In all cases,
\mname{} achieves superior performances more efficiently,
suggesting the effectiveness and robustness of our model.

\section{Conclusion} \label{conclusion}
In this paper,
we propose Transformer with Implicit Edges (\mname),
which aims to trace edge semantics in an edge-free manner
and introduces abstract particles to simulate domains of different complexity and materials,
Our experimental results show the effectiveness and efficiency of our edge-free structure.
The abstract particles enable \mname~to capture material-specific semantics,
achieving robust performances on complex generalization domains.
Finally,
\mname~makes a successful attempt to hybrid GNN and Transformer into physics simulation
and achieve superior performances over existing methods,
showing the potential abilities of implicitly modeling edges in physics simulations.

\noindent
\textbf{Acknowledgements.}
This study is supported under the RIE2020 Industry Alignment Fund Industry Collaboration Projects (IAF-ICP) Funding Initiative, as well as cash and in-kind contribution from the industry partner(s). It is also supported by Singapore MOE AcRF Tier 2 (MOE-T2EP20221-0011) and Shanghai AI Laboratory. 

\clearpage
\bibliographystyle{splncs04}
\bibliography{egbib}

\newpage
\appendix
% \title{Supplementary Material of Transformer with Implicit Edges for Particle-based Physics Simulation} % Replace with your title
% \maketitle

\section{Model Details}
\subsection{Decomposing GNN} \label{app:dec_gnn}
In the following,
we show the detailed deduction of implicitly modeling edges in \mname.
When omitting the normalization, bias, and activation,
we can update the edge propagation function implemented by MLPs in GNN by
\begin{eqnarray}
	\ve^{(l+1)}_{ij}	&	=	&	W^{(l)}\left[\vv^{(l)}_i; \vv^{(l)}_{j}; \ve^{(l)}_{ij}\right],
\end{eqnarray}
where $W^{(l)} \in \mathbb{R}^{d \times 3d}$ is the parameter for MLPs,
and $[\cdot;\cdot]$ denotes the concatenation.
By splitting $W^{(l)}$ into 3 different square blocks $W^{(l)}=[W^{(l)}_r, W^{(l)}_s, W^{(l)}_m]$ and expanding the edge embeddings,
we have
\begin{eqnarray}
	\ve^{(l+1)}_{ij}	&	=	&	W^{(l)}_r \vv^{(l)}_{i} + W^{(l)}_s \vv^{(l)}_{j} + W^{(l)}_m \ve^{(l)}_{ij} \\
					&	=	&	\left(W^{(l)}_r \vv^{(l)}_{i} + W^{(l)}_m W^{(l-1)}_r \vv^{(l-1)}_i\right) \\
					&		&	+ \left(W^{(l)}_s \vv^{(l)}_{j}+W^{(l)}_m W^{(l-1)}_s \vv^{(l-1)}_j\right) \\
					&		&	+ W^{(l)}_m W^{(l-1)}_m \ve^{(l-1)}_{ij}\\
					&	=	&	\left(W^{(l)}_r \vv^{(l)}_{i} + \sum_{u=1}^{l} \left(\prod_{k=0}^{u-1} W_m^{(l-k)}\right) W_r^{(l-u)} \vv_{i}^{(l-u)} \right) \\
					&		&	+\left(W^{(l)}_s \vv^{(l)}_{j} + \sum_{u=1}^{l} \left(\prod_{k=0}^{u-1} W_m^{(l-k)}\right) W_s^{(l-u)} \vv_{j}^{(l-u)} \right), \label{func:e_split}
\end{eqnarray}
where we assume $\vv_i^{(0)}=\vx_i$,
and $[W^{(0)}_r, W^{(0)}_s] = W^{(0)}$ are the parameters for the edge initialization function $f^{\mathrm{enc}}_E(\cdot)$.
Assuming $\vr^{(l)}_i=W^{(l)}_r \vv^{(l)}_i + W^{(l)}_m \vr^{(l-1)}_i$
and $\vs^{(l)}_j=W^{(l)}_s \vv^{(l)}_j + W^{(l)}_m \vs^{(l-1)}_j$,
we can further simplify Eq.\ref{func:e_split} by
\begin{eqnarray}
	\ve^{(l+1)}_{ij}	&	=	&	\left(W^{(l)}_r \vv^{(l)}_i + W^{(l)}_m \vr^{(l-1)}_i\right) + \left(W^{(l)}_s \vv^{(l)}_j + W^{(l)}_m \vs^{(l-1)}_j\right) \\
					&	=	&	\vr^{(l)}_i + \vs^{(l)}_j,
\end{eqnarray}
where we assume $\vr^{(0)}_i=W^{(0)}_r \vv^{(0)}_i$ and $\vs^{(0)}_j=W^{(0)}_s \vv^{(0)}_j$,
$l \in \{1, 2, \cdots, L\}$ is the index of the block.

By combining \Eqref{func:e_split} into the self-attention formula in Transformer \cite{vaswani2017attention}, we have:
\begin{eqnarray}
	\omega^\prime_{ij}		&	=	&	(W^{(l)}_Q \vv^{(l)}_i)^\top (\vr^{(l)}_i + \vs^{(l)}_j) \label{func:attn_rs}\\
							&	=	&	(W^{(l)}_Q \vv^{(l)}_i)^\top \vr^{(l)}_i + (W^{(l)}_Q \vv^{(l)}_i)^\top \vs^{(l)}_j,  \label{func:v2v} \\
	\hat{\omega}^\prime_{ij}		&	=	&	\softmax\left(\frac{\omega^\prime_{ij}}{\sqrt{d}}\right), \\
	\vv^{(l+1)}_i			&	=	&	\sum_j \hat{\omega}^\prime_{ij} \cdot (\vr^{(l)}_i+\vs^{(l)}_j) \label{func:vrps}\\
							&	=	&	\sum_j \hat{\omega}^\prime_{ij} \cdot \vr^{(l)}_i + \sum_j \hat{\omega}^\prime_{ij} \cdot \vs^{(l)}_j \\
							&	=	&	\vr^{(l)}_i + \sum_j \hat{\omega}^\prime_{ij} \cdot \vs^{(l)}_j. \label{func:ve}
\end{eqnarray}

\subsection{Normalization Effects} \label{app:dec_meanstd}
Given \Eqref{func:e_split},
we further modify the self-attention in \Eqref{func:v2v} and \Eqref{func:ve}
to include the effects of normalization in GNN for edges.
Since GNN-based methods usually incorporate LayerNorm \cite{DBLP:journals/corr/BaKH16}
in their network architectures that computes the mean and std of edge features
to improve their performance and training speed,
we propose to apply the effects of normalization for each edge in GNN to our model.
For the mean $\mu^{(l)}_{ij}$ and std $\sigma^{(l)}_{ij}$ of each interaction between particle $i$ and $j$,
we can compute them from receiver token $\vr^{(l)}_{i}$ and sender token $\vs^{(l)}_{j}$ by
\begin{eqnarray}
	\mu^{(l)}_{ij}	&	=	&	\frac{1}{d} \sum_k \left(r^{(l)}_{ik} + s^{(l)}_{jk} \right) \\
			&	=	&	\frac{1}{d} \left(\sum_k r^{(l)}_{ik} + \sum_k s^{(l)}_{jk} \right) \\
			&	=	&	\mu^{(l)}_{r_i} + \mu^{(l)}_{s_j},
\end{eqnarray}
\begin{eqnarray}
	\left(\sigma^{(l)}_{ij}\right)^2	&	=	&	\frac{1}{d} \sum_k \left(r^{(l)}_{ik} + s^{(l)}_{jk} -\mu^{(l)}_{ij}\right)^2 \\
				&	=	&	\frac{1}{d} \sum_k \left((r^{(l)}_{ik})^2 + (s^{(l)}_{jk})^2 + (\mu_{ij}^{(l)})^2 + 2r^{(l)}_{ik} s^{(l)}_{jk} -2\mu_{ij}^{(l)}(r^{(l)}_{ik} + s^{(l)}_{jk})\right) \\
				&	=	&	\frac{1}{d} \left((\vr^{(l)}_{i})^\top\vr^{(l)}_{i} + (\vs^{(l)}_{j})^\top\vs^{(l)}_{j}\right) + (\mu^{(l)}_{ij})^2 + \frac{2}{d}(\vr^{(l)}_{i})^\top \vs^{(l)}_{j} -2(\mu^{(l)}_{ij})^2 \\
				&	=	&	\frac{1}{d} (\vr^{(l)}_{i})^\top\vr^{(l)}_{i} + \frac{1}{d} (\vs^{(l)}_{j})^\top\vs^{(l)}_{j} + \frac{2}{d}(\vr^{(l)}_{i})^\top \vs^{(l)}_{j} -(\mu^{(l)}_{ij})^2\\
				&	=	&	\frac{1}{d} (\vr^{(l)}_{i})^\top\vr^{(l)}_{i} + \frac{1}{d} (\vs^{(l)}_{j})^\top\vs^{(l)}_{j} + \frac{2}{d}(\vr^{(l)}_{i})^\top \vs^{(l)}_{j} -(\mu^{(l)}_{r_i} + \mu^{(l)}_{s_j})^2
\end{eqnarray}
where $r^{(l)}_{ik}$ and $s^{(l)}_{jk}$ are respectively the $k$-th element
in $\vr^{(l)}_{i}$ and $\vs^{(l)}_{j}$,
$\mu^{(l)}_{r_i}$ and $\mu^{(l)}_{s_j}$ are respectively
the mean of receiver token $\vr^{(l)}_{i}$ and sender token $\vs^{(l)}_{j}$ after $l$-th block.
Hence,
by replacing $\vr^{(l)}_i + \vs^{(l)}_j$ in \Eqref{func:attn_rs} and \Eqref{func:vrps}
by $\frac{\vr^{(l)}_i + \vs^{(l)}_j-\mu^{(l)}_{ij}}{\sigma^{(l)}_{ij}}$,
we have
\begin{eqnarray}
	\omega^{\prime\prime}_{ij}	&	=	&	\frac{(W^{(l)}_Q\vv^{(l)}_i)^\top (\vr^{(l)}_i-\mu^{(l)}_{r_i}) + (W^{(l)}_Q \vv^{(l)}_i)^\top (\vs^{(l)}_j-\mu^{(l)}_{s_j})}{\sigma_{ij}^{(l)}}, \\
	\hat{\omega}^{\prime\prime}_{ij}		&	=	&	\softmax\left(\frac{\omega^{\prime\prime}_{ij}}{\sqrt{d}}\right), \\
	\vv^{(l+1)}_i	&	=	&	\sum_j \hat{\omega}^{\prime\prime}_{ij} \cdot \frac{\vr^{(l)}_i-\mu^{(l)}_{r_i}}{\sigma_{ij}^{(l)}} + \sum_j \hat{\omega}^{\prime\prime}_{ij} \cdot \frac{\vs^{(l)}_j - \mu^{(l)}_{s_j}}{\sigma_{ij}^{(l)}}, \label{func:agg_tie}
\end{eqnarray}
When it comes to the scaling and shifting parameters in LayerNorm \cite{DBLP:journals/corr/BaKH16},
we add them into \Eqref{func:agg_tie} to better resemble the normalization effects.

\section{Experiment Details}
\subsection{Implementation Details}\label{app:impl}
\noindent
\textbf{Inputs and outputs details}.
For \emph{FluidFall}, \emph{FluidShake}, and \emph{BoxBath},
we only use particles' states at time $t$ as inputs
and output the velocities at time $t+1$.
For \emph{RiceGrip},
we concatenate particles states from $t-2$ to $t$ as inputs
and output 6-dim vector for the velocity of the current observed position and the resting position.
For \emph{BoxBath},
we output 7-dim vectors,
where 3 dimensions for the predicted velocities,
and 4 dimensions for rotation constrains.
The rotation constraints, which predict the rotation velocities,
are applied only on rigid particles,
which is the same as mentioned in DPI-Net \cite{li2019learning}.
All states of particles, such as the positions and velocities,
are first normalized by mean and standard deviations calculated on corresponding training set
before they are fed into the models.

\noindent
\textbf{Training}.
We train four models independently on four domains,
with 5 epochs on \emph{FluidShake} and \emph{BoxBath},
13 epochs on \emph{FluidFall},
and 20 epochs on \emph{RiceGrip}.
For common settings,
we adopt Adam optimizer with an initial learning rate of 0.0008,
which has a decreasing factor of 0.8 when the validation loss stops to decrease after 3 epochs.
The batch size is set to 16 on all domains.
All models are trained and tested on V100 for all experiments,
with no augmentation involved.

\noindent
\textbf{Baseline details}.
For fair comparison,
the following settings are the same with \mname:
inputs for models, 
number of training epochs on different domains,
learning rate schedules,
and training loss on velocities.
Hyper-parameters for baselines are first chosen the same as their original papers, and then fine-tuned within a small range of changes.
For example,
in terms of the batch size, 16 works better for DPI-Net than the original settings.

\subsection{Data Generation}
\begin{table}[t]
	\caption{
		Details of generalization settings.
		We list the number of particles in both training domains and generalization domains.
		The lists of numbers in \emph{L-FluidShake} and \emph{L-RiceGrip} are the range of particles,
		while the number of rigid and fluid particles in generalized \emph{BoxBath} are listed separately.}
	\label{tbl:gen_details}
	\setlength{\tabcolsep}{4.0pt}
	\begin{center}
	\small
	\begin{tabular}{l|ll}
	\toprule
	\multicolumn{1}{c}{\bf Domains}	& \multicolumn{1}{|c}{\bf Training Settings}	& \multicolumn{1}{c}{\bf Generalization Settings}
	\\ \midrule
	L-FluidShake         			& [450, 627] 									& [720, 1368] \\ 
	L-RiceGrip						& [570, 980]									& [1062, 1642]\\
	Lfluid-BoxBath					& Fluid: 960. Rigid: 64							& Fluid: 1280. Rigid 64\\
	L-BoxBath						& Fluid: 960. Rigid: 64							& Fluid: 960. Rigid: 125\\
	BunnyBath						& Fluid: 960. Rigid: 64							& Fluid: 960. Rigid: 41\\
	BallBath						& Fluid: 960. Rigid: 64 							& Fluid: 960. Rigid: 136\\
	\bottomrule
	\end{tabular}
	\end{center}
	\end{table}
	
\noindent
\textbf{Basic Domains.}
We use the same setting for our datasets as mentioned in previous work \cite{li2019learning}.
\emph{FluidFall} contains two fluid droplets with different sizes.
The sizes for droplets are randomly generated with one droplet larger than the other.
Positions and viscosity for droplets are randomly initialized.
This domain contains 189 particles with 121 frames for each rollout.
There are 2700 rollouts in training set and 300 rollouts in validation set.
\emph{FluidShake} simulates the water in a moving box.
The speed of the box is randomly generated at each timestamp.
In addition,
the size of the box and the number of particles are various for different rollouts.
In basic training and validation sets,
the number of particles varies from 450 to 627.
This domain has 301 frames for each rollout.
There are 1800 rollouts in training set and 200 rollouts in validation set.
\emph{RiceGrip} contains two grippers and a sticky rice.
The grippers' positions and orientations are randomly initialized.
The number of particles for rice varies from 570 to 980 with 41 frames for each rollout in training and validation sets.
There are 4500 rollouts in training set and 500 rollouts in validation set.
\emph{BoxBath} simulates a rigid cube washed by water in a fixed container.
The initial positions of fluid block and rigid cube are randomly initialized.
This domain contains 960 fluid particles and 64 rigid particles with 151 frames for each rollout.
There are 2700 rollouts in training set and 300 rollouts in validation set.

\noindent
\textbf{Generalization Domains.}
We release the details of generalization settings in Table \ref{tbl:gen_details}.
We add more particles for \emph{FluidShake} and \emph{RiceGrip},
which we refer to as \emph{L-FluidShake} and \emph{L-RiceGrip} respectively.
The \emph{L-FluidShake} includes 720 to 1368 particles,
while \emph{L-RiceGrip} contains 1062 to 1642 particles.
On \emph{BoxBath},
we enlarge the fluid block and change the size and shape of rigid object.
Specifically,
we add more fluid particles in \emph{Lfluid-BoxBath} to 1280 fluid particles,
while we enlarge the rigid cube in \emph{L-BoxBath} to 125 particles.
We also change the shape of the rigid object into
ball and bunny,
which we refer to \emph{BallBox} and \emph{BunnyBath} respectively.
The number of test rollouts and the number of frames for each rollout
are the same as the corresponding basic domains.

\subsection{Rendered Rollouts}
We visualize some rollouts on \emph{BoxBath} and its generalized domains,
which are complex domains with multi-materials interactions.
We simplify \emph{BoxBath} into 5 key steps:
1. the rigid object flooded by fluid hits the wall;
2. the rigid object is thrown into the air;
3. the rigid object falls into the fluid;
4. the fluid, after hitting the wall, pushes the rigid object;
5. the rigid object slows down and stops moving.
The visual results and analysis are shown in the following.
Our \mname+ achieves more faithful rollouts on all the domains,
suggesting the effectiveness of our implicitly modeled edges and the abstract particles.

\begin{figure}[t]
	\begin{center}
		\includegraphics[width=120mm]{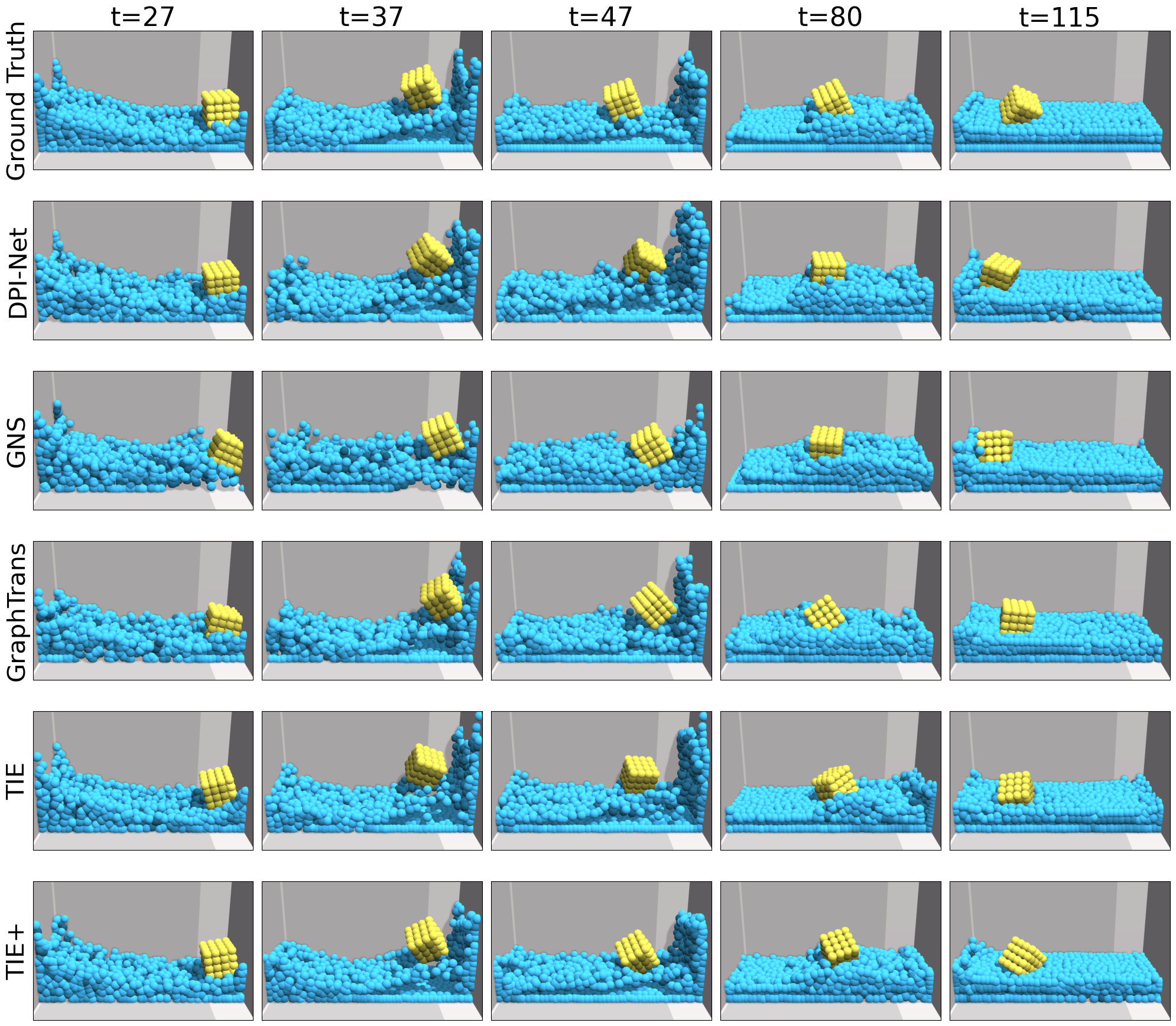}
		\caption{
		\small{
		Qualitative results on \emph{BoxBath}.
		For step 2 and 3,
		the rigid cube is flooded into the air and pushed away from the right wall.
		Our \mname+ is able to achieve more faithful rollouts even with more accurate rotation angles,
		and the cube is pushed far away enough from the right wall.
		For step 5,
		when the cube slows down and finally stops,
		the position of the rigid cube predicted by our \mname+ is closer to the ground truth.
		For the simulations of the fluid particles,
		our model achieves more vivid results compared with other models.
		For example,
		the surface of water is smooth and is closer to the ground truth.
		}}
		\label{fig:boxbath}
	\end{center}
\end{figure}

\begin{figure}[t]
	\begin{center}
		\includegraphics[width=120mm]{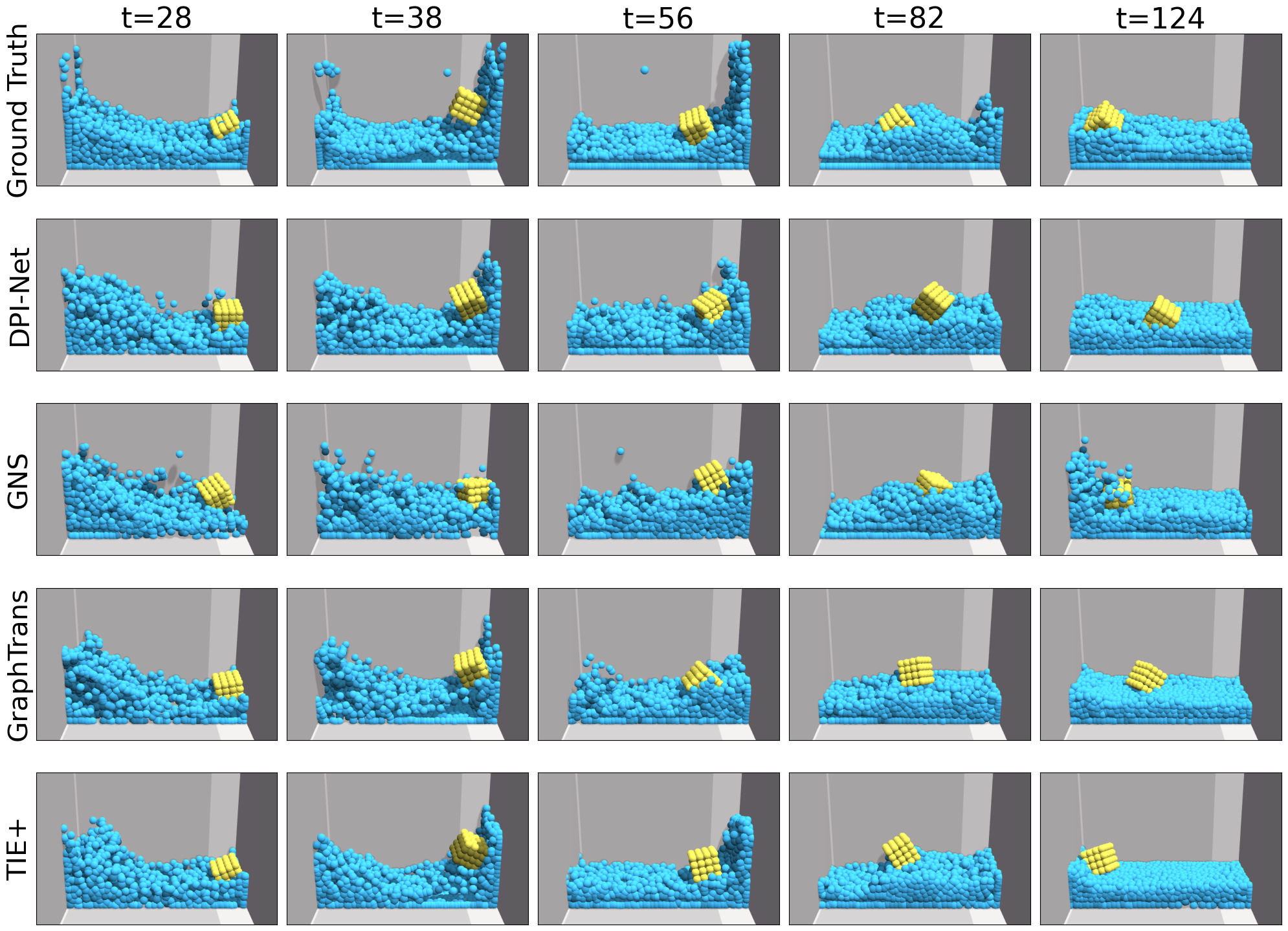}
		\caption{
		\small{
		Qualitative results on \emph{Lfluid-BoxBath},
		where we add more fluid particles.
		When focusing on the rigid cube,
		the positions and rotation angles achieved by our \mname+
		are much closer to the ground truth.
		When focusing on the fluid particles,
		our \mname+ predicts more vivid wave,
		which floods the box and pushes it towards left in a more faithful manner.
		}}
		\label{fig:lfluidboxbath}
	\end{center}
\end{figure}

\begin{figure}[t]
	\begin{center}
		\includegraphics[width=120mm]{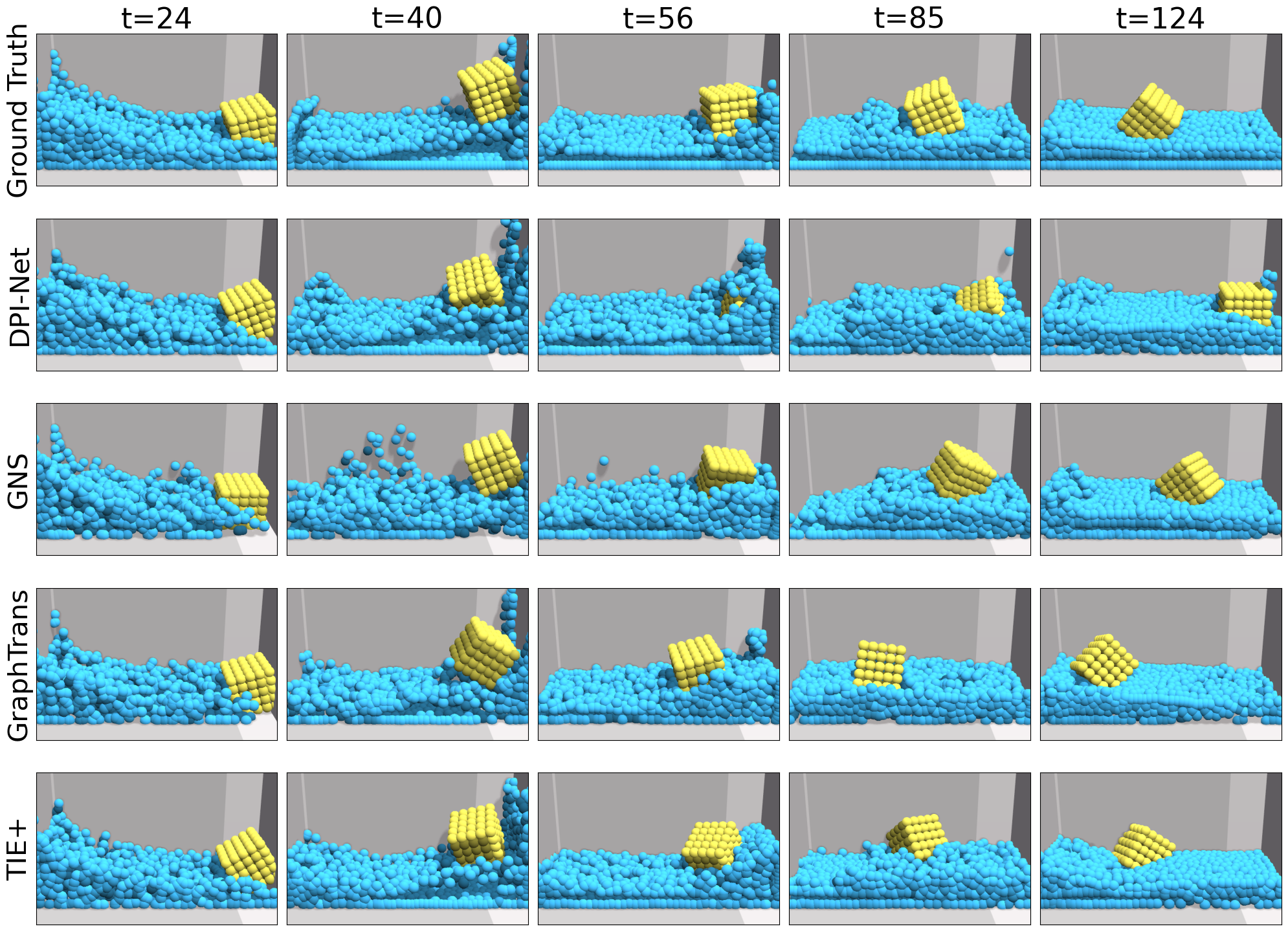}
		\caption{
		\small{
		Qualitative results on \emph{L-BoxBath},
		where we enlarge the rigid cube.
		Notice that the rigid cubes predicted by DPI-Net and GNS tends to rotate
		and move in the same place,
		DPI-Net and GNS have more difficulties in being generalized to this domain.
		The rigid cube predicted by GraphTrans is overly pushed by the wave at $t=56$,
		while does not fully interact with the left wall to bounce back at $t=124$.
		For both the fluid part and the rigid part,
		\mname+ still predicts more faithful results.
		}}
		\label{fig:lboxbath}
	\end{center}
\end{figure}

\begin{figure}[t]
	\begin{center}
		\includegraphics[width=120mm]{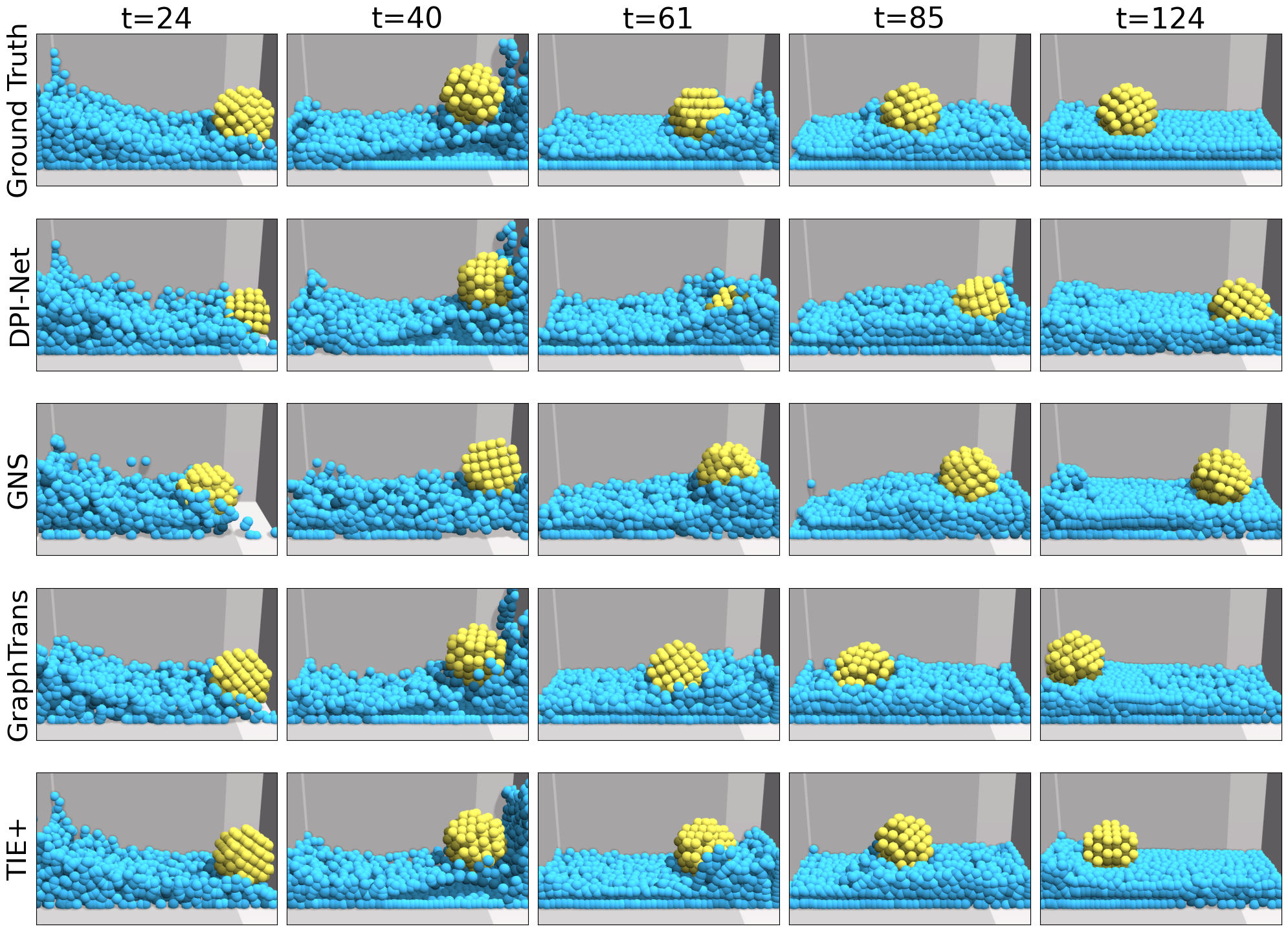}
		\caption{
		\small{
		Qualitative results on \emph{BallBath},
		where we change the cube into ball.
		The balls predicted by both DPI-Net and GNS rotate and move in the same place till the end,
		while GraphTrans has difficulties predicting the positions of the ball.
		\mname+ still achieves faithful rollout.
		}}
		\label{fig:ballbath}
	\end{center}
\end{figure}

\begin{figure}[t]
	\begin{center}
		\includegraphics[width=120mm]{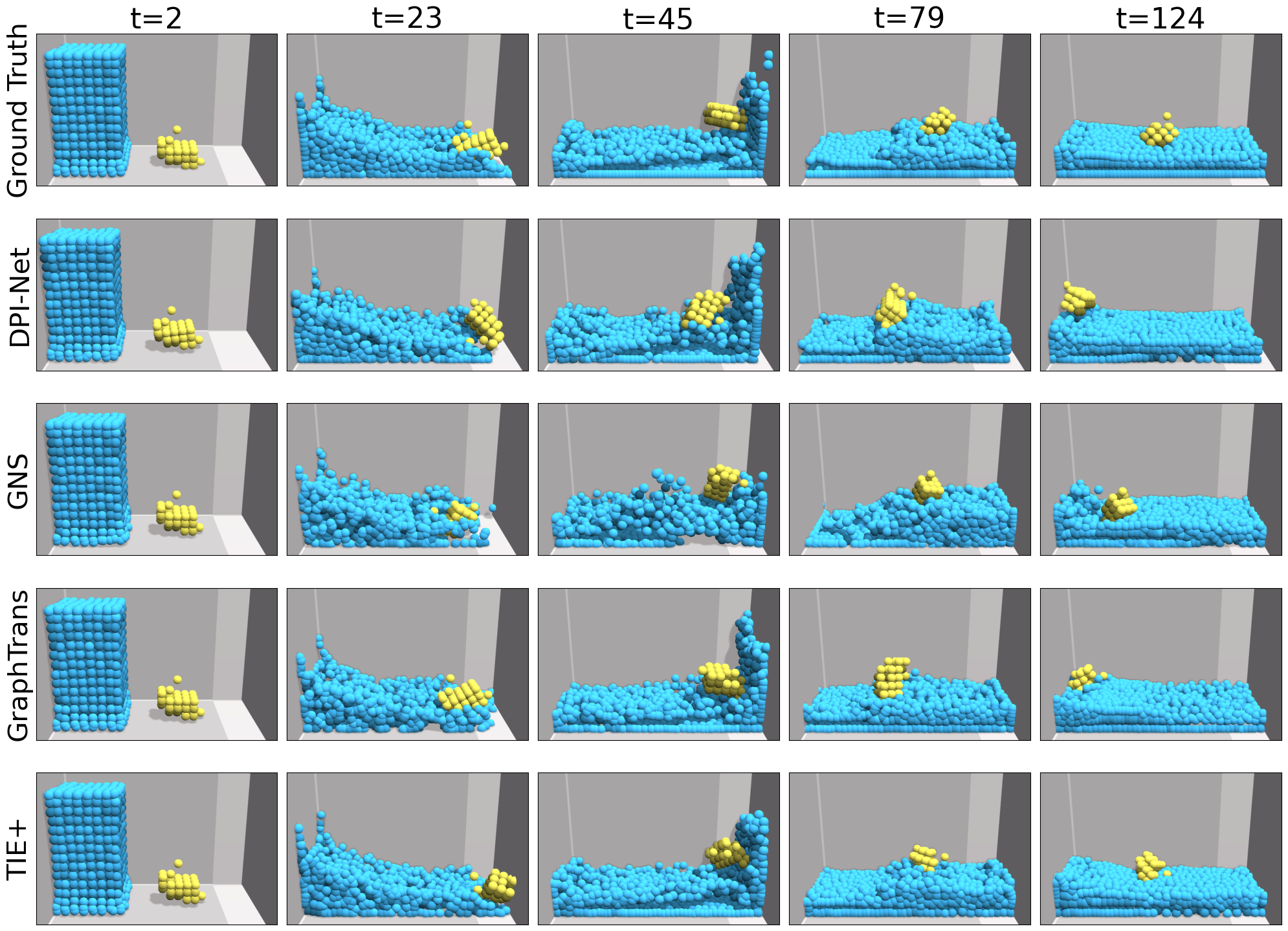}
		\caption{
		\small{
		Qualitative results on \emph{BunnyBath},
		where we change the cube into bunny with more complex surfaces.
		\mname+ achieves more faithful rollout.
		At time $t=2$,
		we show the shape of the bunny,
		which has complex surfaces.
		At time $t=45$,
		while \mname+ is able to rollout vivd dynamics of fluid particles,
		\mname+ can predict closer positions of the bunny,
		which is rotating and flying in the air.
		}}
		\label{fig:bunnybath}
	\end{center}
\end{figure}

\end{document}